\pdfoutput=1

\documentclass[11pt]{article}

\usepackage[preprint]{acl}

\usepackage{times}
\usepackage{latexsym}

\usepackage[T1]{fontenc}

\usepackage[utf8]{inputenc}

\usepackage{microtype}

\usepackage{inconsolata}

\usepackage{graphicx}
\usepackage{pifont}
\usepackage{booktabs}
\newcommand{\cmark}{\ding{51}}
\newcommand{\xmark}{\ding{55}}
\usepackage{adjustbox}
\title{Data-Centric Improvements for Enhancing Multi-Modal Understanding in Spoken Conversation Modeling}

\author{Maximillian Chen\thanks{Work done during an internship at Google.} \\
  Columbia University and Google \\
  \texttt{maxchen@cs.columbia.edu} \\\And
  Ruoxi Sun \\
  Google \\
  \texttt{ruoxis@google.com} \\\And
  Sercan \"{O}. Ar{\i}k \\
  Google \\
  \texttt{soarik@google.com} \\}

\begin{document}
\maketitle
\begin{abstract}
Conversational assistants are increasingly popular across diverse real-world applications, highlighting the need for advanced multimodal speech modeling. Speech, as a natural mode of communication, encodes rich user-specific characteristics such as speaking rate and pitch, making it critical for effective interaction. Our work introduces a data-centric customization approach for efficiently enhancing multimodal understanding in conversational speech modeling. Central to our contributions is a novel multi-task learning paradigm that involves designing auxiliary tasks to utilize a small amount of speech data. Our approach achieves state-of-the-art performance on the Spoken-SQuAD benchmark, using only 10\% of the training data with open-weight models, establishing a robust and efficient framework for audio-centric conversational modeling. We also introduce ASK-QA, the first dataset for multi-turn spoken dialogue with ambiguous user requests and dynamic evaluation inputs. 
\end{abstract}

\section{Introduction}
Real-world adoption of intelligent multimodal conversational agents has progressed quickly in recent years due to the impressive capabilities of Large Language Models (LLMs). 
However, despite numerous applications, including smart home systems, contact centers, customer support/service, personalized education etc. ~\citep{hemphill1990atis, khatri2018alexa,li2017acoustic,von2013duolingo,fatima2024accuracy,li2023speech,zhenglmsys}, there has not been the same rapid progress in adapting Multimodal LLMs (MLLMs) to spoken contexts due to several fundamental challenges. 

\begin{figure}
    \centering
    \includegraphics[width=0.8\linewidth]{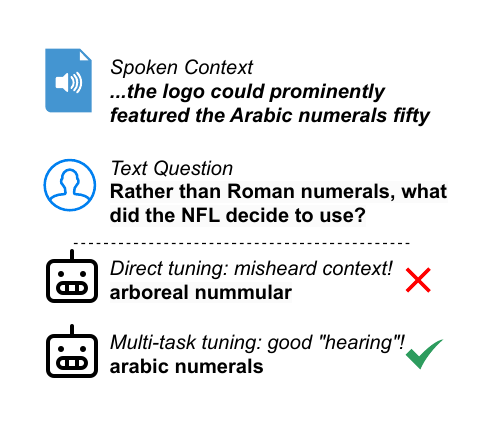}
    \vspace{-0.6cm}
    \caption{Automatic speech recognition is a necessary \textit{implicit} skill for MLLM in end-to-end spoken question answering. We propose a multi-task learning approach which explicitly teaches these skills, as exemplified by this QA pair from Spoken-SQuAD.}
    \label{fig:fig1}
\end{figure}

Speech data constitute high-dimensional signals that are difficult to model even for frontier models (e.g., Whisper-based models are limited to 30 seconds;~\citet{chu2024qwen2,radford2023robust}) and Gemini 1.5 represents 1 second of audio using 25 tokens\footnote{\url{https://ai.google.dev/gemini-api/docs/audio?lang=python}}). These are temporal signals which include acoustic phenomena (e.g. background noise~\citet{mehrish2023review}) and important paralinguistic information such as speaking rates or pitch~\cite{hirschberg1993pitch,bhattacharya2023capturing}. 
Performance on speech understanding tasks is thus greatly affected by the ability to robustly comprehend the semantic contents of the input  speech~\cite{li2017acoustic}, as illustrated in Fig.~\ref{fig:fig1}. This is further complicated by the long-standing issue of models overfitting to individual speakers~\cite{jung2018avoiding,wang2020spoken}. 
These can be viewed as a shortcoming of insufficient training data coverage~\cite{yang2024audiobox}. However, large-scale speech data collection is notoriously difficult due to privacy concerns~\cite{nautsch2019preserving, qian2018towards}. 

Despite the difficulty of large-scale collection, task-specific data is increasingly the most effective approach to guarantee use-case customization for state-of-the-art MLLMs like Gemini or GPT~\cite{team2023gemini,brown2020language}. These models are closed-source, but offer commercial tuning APIs, which typically do not permit modifications to the model or learning objective. Even with smaller open-weight models, it can still be computationally intractable to iterate on architectures and train from scratch due to the expensive compute resource demands~\cite{OLMo}.
These motivate the design of efficient data-centric methods~\cite{seedat2022dc} which maximize models' ability to overcome the aforementioned challenges of long speech understanding reliably.

In this work, we take a data-centric perspective towards addressing the varied challenges of adapting multimodal LLMs for speech. Our contributions can be summarized as follows:
\begin{itemize}
    \setlength\itemsep{0em}
    \item We bring a multi-task learning paradigm to improve speech understanding implemented via a simple but effective data-centric approach. Rather than using additional datasets, we instead design auxiliary tasks to maximize cross-modal learning from a fixed set of recorded speech.
    \item We propose \textbf{A}mbiguous \textbf{S}po\textbf{k}en Conversational \textbf{Q}uestion \textbf{A}nswering (ASK-QA), a novel dataset which combines the challenges of multimodal speech modeling and mixed-initiative interaction. ASK-QA features contextually ambiguous questions along with long multi-turn speech and diverse accents, speaking rates, and pitches. 
    \item We validate the proposed data-centric approach on three spoken question answering (SQA) corpora: ASK-QA, Spoken-SQuAD, and SD-QA, representing various combinations for whether input questions and knowledge context are represented as text or speech. Our approach applied even to open-weight models is able to outperform the existing state-of-the-art on Spoken-SQuAD using only 10\% of the available training data.\footnote{Code and data forthcoming.}
\end{itemize}
\section{Data-Centric Multi-Task Learning for Cross-Modal Understanding}
\label{sec:recipe}
We consider the setting of customizing an MLLM for use in request-based end-to-end speech modeling, similarly to \citet{shih2023gsqa}.
An MLLM is provided as input an audio recording and textual context. The backbone of many MLLM architectures is a textual decoder-only LLM~\cite{liu2024visual}, so the textual context usually contains an instruction. These settings involve reasoning about some contextual knowledge and conversation history. The model aims to provide a correct answer to a target question (i.e. the last conversation turn).  Different applications may involve spoken conversations about written information (e.g. document-grounded QA), or written conversations about spoken information (e.g. meeting summarization). 

Tuning MLLMs with cross-entropy loss is advantageous as it can be used to unify diverse tasks as a single text-to-text objective~\cite{raffel2020exploring}. Many recent studies find that multi-task learning~\cite{caruana1997multitask} using \textit{additional datasets} greatly improves downstream task performance~\cite{aghajanyan2021muppet,padmakumar2022exploring,chen2023pre}. 
\textit{Here, we instead design auxiliary tasks within the same dataset to maximize cross-modal learning gains from a fixed set of audio recordings for a target dataset}. We break down our problem into three intermediate goals: 1) correctly representing the spoken context, 2) learning to reason across all input modalities, and 3) coherently producing the correct answer. 

\paragraph{1) Listening Comprehension} is an auxiliary task to help the SLM ``hear'' the spoken context. It has been consistently reported in traditional cascade-style systems that SQA performance is greatly affected by automatic speech recognition (ASR) errors~\cite{li2018spoken}, and thus we design a task to specifically address this point. The objective is for the MLLM to predict a ground-truth (or pseudo-labeled) audio transcription, given a recording and a task instruction as input.

\paragraph{2) Cross-Modal Commonsense Reasoning}
is an auxiliary task designed to unify the contents of the spoken and textual inputs. We reframe dialogue response selection~\cite{henderson2019training} as a multiple-choice reasoning task~\cite{talmor-etal-2019-commonsenseqa}. The answer options consist of the correct answer (e.g. ``It was recovered a few months later'') and commonsense negative answer choices sampled from other training QA pairs (e.g. ``Do you mean the popular generic name?''), as shown in Table~\ref{askqa_rs}. The objective is to solve this multiple-choice reasoning task by selecting the correct answer given the recording, conversation context, knowledge, answer options, and a task instruction.

\paragraph{3) Response Generation}
is the primary objective of providing a correct answer. The inputs are what is expected to be provided to an MLLM at inference time for SQA: the recording, conversation context, information context, and a task-specific instruction. An example is shown in Table~\ref{askqa_rg}.

These tasks can be fully implemented as modifications to tuning data mixtures. As shown in Sec.~\ref{sec:experiments}, this simple modification is observed to be highly effective in improving an MLLMs' ability to complete downstream tasks, particularly in the limited data regime. We provide implementation details for our approach in Appendix~\ref{sec:multitask_implementation}.

\begin{table}[h]
    \centering
    \scalebox{0.9}{
    \small
    \begin{tabular}{lccc}
        \toprule
         & SD-QA & S-SQuAD & ASK-QA (Ours) \\
        \midrule
        Avg. Audio Len. & 4.8s & 59s & 1m 41s \\
        Speakers/Audio & 1 & 1 & 3 \\
        Knowledge & Text & Speech & Speech \\
        Conversation & Speech & Text & Speech \\
        Unique Voices & 248 & 1 & 64 \\
        Avg. Turns & 2 & 2 & 5.1 \\
        Answer Type & Span & Span & Free-form \\
        Ambiguous & \xmark & \xmark & \cmark \\
        Dynamic Eval & \xmark & \xmark & \cmark \\
        Disfluencies & \xmark & \xmark & \cmark \\
        \bottomrule
    \end{tabular}
    }
    \vspace{-2.5mm}
    \caption{\textbf{Comparison of ASK-QA against existing popular SQA training datasets used for experimentation here}. ASK-QA features ambiguous requests and long audio context. See Appendix~\ref{sec:sdqa_overview} for SD-QA.}
    \label{tab:data_comparison}
    \vspace{-4mm}
\end{table}
\section{Experiments}
\label{sec:experiments}
We evaluate our approach on SQA datasets with different combinations of context modalities (see Table~\ref{tab:data_comparison}). We prompt and fine-tune two closed-source models, Gemini Pro and Gemini Flash, on the Vertex AI platform\footnote{\url{https://cloud.google.com/vertex-ai/generative-ai/docs/models/tune_gemini/audio_tune}}. We also use Speech-Qwen, which we built by pre-training an 17.8M parameter projection layer between a frozen audio encoder (WavLM-Large) and a frozen LLM decoder (Qwen 2.5 7B-Instruct). See Appendix~\ref{sec:adapter_details} for details on Speech-Qwen. Our main findings for ASK-QA and Spoken-SQuAD are reported in this section. Full results and extended details for each experimental setting are reported in Appendix~\ref{sec:extended_experiments}. 

\subsection{ASK-QA: Spoken Knowledge and Multi-Turn Spoken Dialogue}
\label{sec:askqa_experiments}
We develop a novel corpus for speech-based mixed-initiative conversation: \textbf{A}mbiguous \textbf{S}po\textbf{k}en Conversational \textbf{Q}uestion \textbf{A}nswering  (ASK-QA). The contextual inputs for ASK-QA are \textit{fully spoken}.

\begin{figure}[t]
    \centering
    \vspace{-4mm}
    \includegraphics[width=0.8\linewidth]{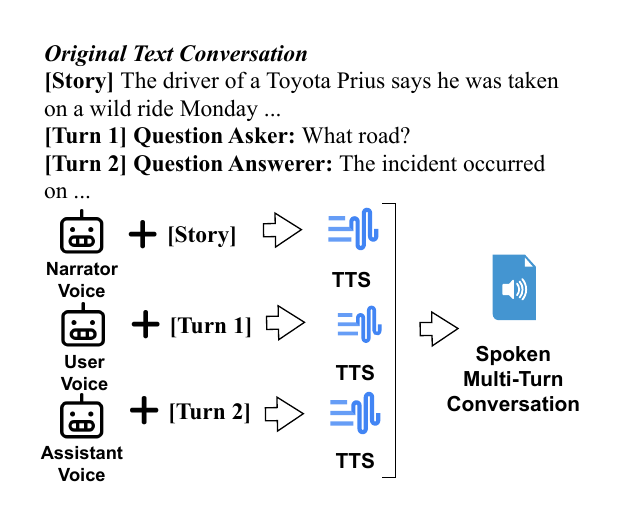}
    \vspace{-0.5cm}
    \caption{\textbf{Simplified summary of the pipeline for constructing ASK-QA.} For each text conversation in Abg-CoQA, we construct three speaker profiles with randomly sampled voices, speaking rates, and pitches. We use TTS to synthesize the story context as a spoken narration, then each individual dialogue turn. The resulting audio files are joined as a single recording.}
    \vspace{-4mm}
    \label{fig:data_generation}
\end{figure}

\paragraph{Dataset Construction:}
\label{sec:speech_generation}
We construct ASK-QA starting from Abg-CoQA~\cite{guo2021abg}, a span-based textual conversational QA task. Given a story as context, each conversation consists of dialogue turns where a user asks questions and an assistant is supposed to provide the correct answer or ask a clarifying question if the user's request is ambiguous.
Our data construction pipeline is summarized in Figure~\ref{fig:data_generation}. 
The data generation process is detailed in Appendix~\ref{sec:askqa_details}. In total, ASK-QA contains 221.8h of speech. The training, validation, and test sets contain 5,985, 500, and 1,345 conversations.

\paragraph{Evaluation:}
Following recent work~\citep{chen2024learning,risch-etal-2021-semantic}, we apply embedding-based semantic similarity~\cite{reimers2019sentence} to allow for flexible free-form QA evaluation. We apply this metric to a standard single-turn setting as well as a novel multi-turn setting which combines TTS with the dynamic input evaluation for Abg-CoQA in \citet{chen2024learning}. See details in Appendix~\ref{sec:evaluation_details}.

\begin{figure}[!t]
    \centering
    \vspace{-4mm}
    \includegraphics[width=0.99\linewidth]{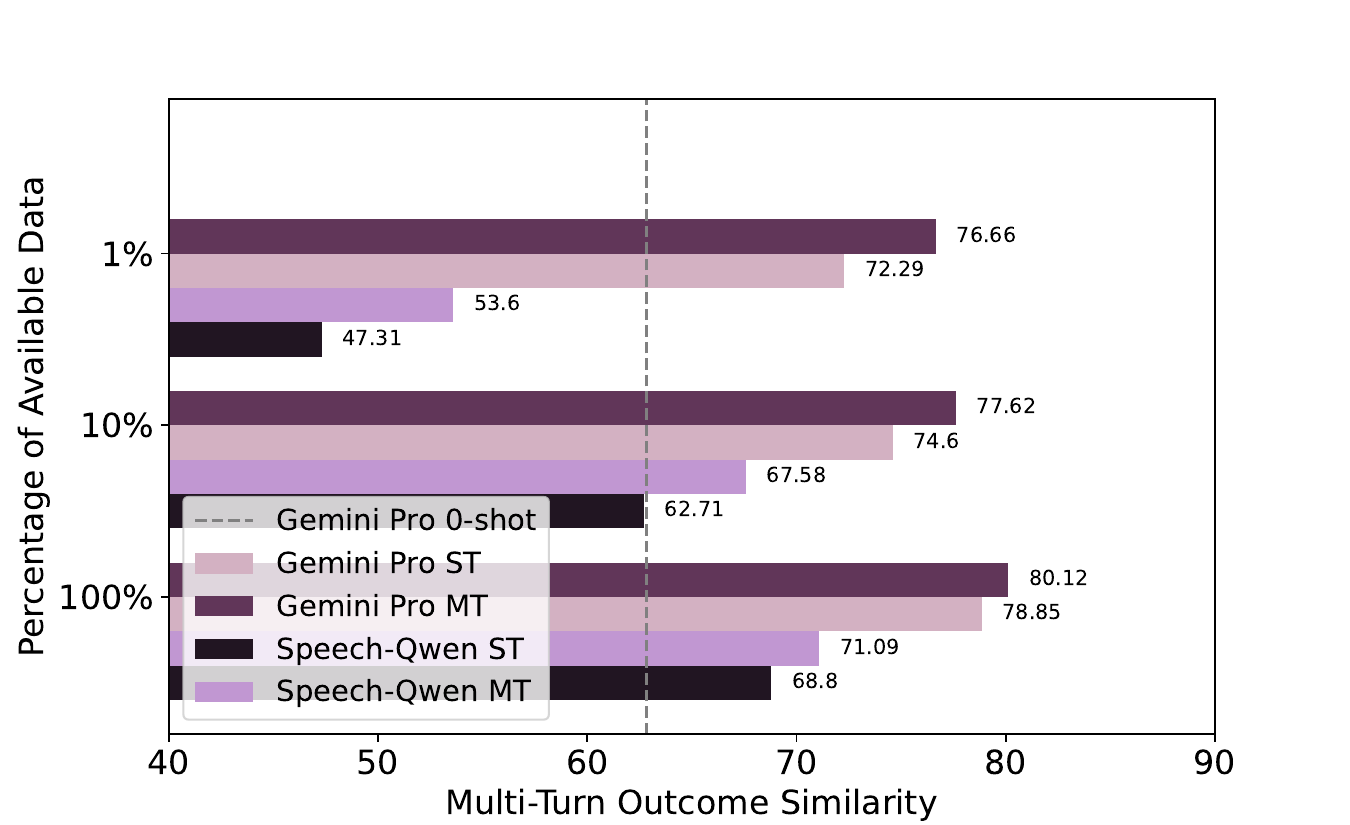}
    \caption{Multi-task (MT) learning improves upon Single-task (ST) fine-tuning with both Gemini and Speech-Qwen on ASK-QA's multi-turn evaluation.}
    \vspace{-3mm}
    \label{fig:askqa_multiturn_results}
\end{figure}

\paragraph{Findings:}
We benchmark end-to-end performance on ASK-QA in Table~\ref{tab:askqa_results} using the multi-task approach described in Section~\ref{sec:recipe} and baseline single-task tuning (which represents standard end-to-end speech-to-text modeling~\cite{shih2023gsqa}).
The listening comprehension sub-task separately models the story and conversation transcripts, inspired by speaker diarization~\cite{anguera2012speaker,gu2021mpc,yu2022speaker}. 
With Speech-Qwen, we see as much as 13.3\% relative improvement over standard fine-tuning depending on the amount of available data on trajectory-level similarity in Figure~\ref{fig:askqa_multiturn_results}. Surprisingly, with Gemini Pro, we also see relative improvements of 5.7\% with 1\% of the available training data and 1.6\% when using full data,
despite frontier MLLMs already having large-scale multi-modal pre-training and the full ASK-QA corpus containing large-scale, in-distribution data (over 200 hours). This finding is significant because it \textit{specifically indicates that the MLLM is better learning to model the available speech data.} It is well-documented that 1) high benchmark scores achieved by frontier LLMs on older corpora may be confounded by data contamination~\cite{roberts2023cutoff, qian-etal-2024-varbench}, and 2) several studies demonstrate the efficacy of direct fine-tuning given abundant data~\cite{sharma2024critical,yu2024lions}. Since ASK-QA is newly synthesized, Gemini cannot have been trained on this exact version of the data. This accurately highlights the difference between direct single-task tuning and multi-task tuning with our proposed auxiliary tasks. The improvements with full data indicate the applicability of the approach for scaled data.
\subsection{Spoken-SQuAD: Spoken Knowledge and Textual Questions}
\label{sec:squad_overview}
Spoken-SQuAD~\citep{li2018spoken} is a speech version of SQuAD~\cite{rajpurkar2016squad}.
Rather than span-based classification, we solve the task using our end-to-end generative approach. Each instance has a textual question and spoken knowledge.

\begin{figure}[t]
    \centering
    \vspace{-4mm}
    \includegraphics[width=0.99\linewidth]{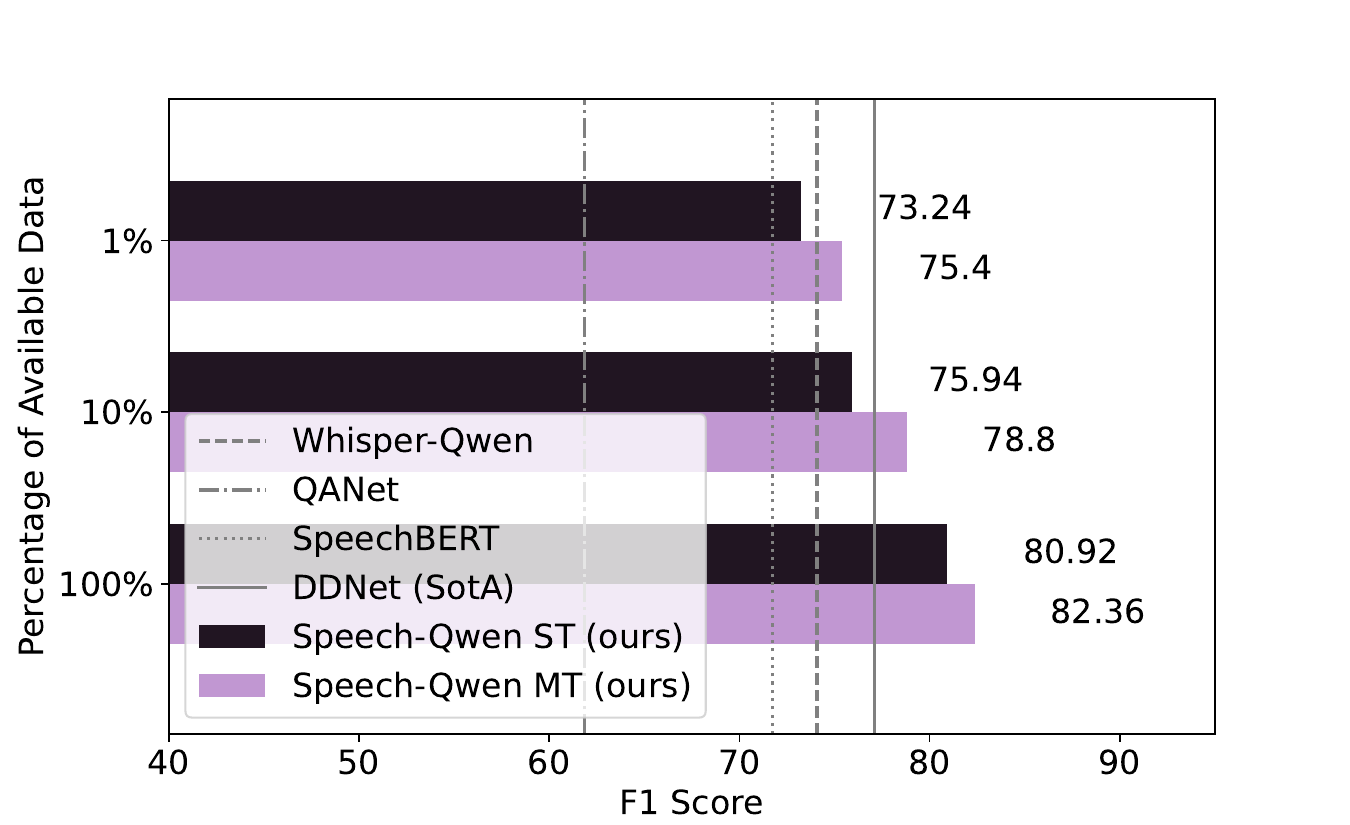}
    \caption{Our multi-task approach applied to Speech-Qwen outperforms the state-of-the-art approach on Spoken-SQuAD using only 10\% of the available data.}
    \vspace{-3mm}
    \label{fig:squad_mainresults}
\end{figure}

\paragraph{Findings:}
In Figure~\ref{fig:squad_mainresults}, we benchmark our multi-task learning approach against single-task tuning via Speech-Qwen. Our performance is evaluated in terms of exact match and F1 score using the SQuAD evaluator. Our approach, applied to an open-weight model like Speech-Qwen, outperforms the existing state-of-the-art model proposed in \citet{you-etal-2022-end} using just 10\% of the available training data, indicating that it is \textit{highly efficient and effective for cross-modal learning}. Our expanded results which include an additional MLLM are shown in Table~\ref{tab:squad_results}. We see similar trends (up to 52.8\% improvement given limited data) on the multi-lingual SD-QA corpus in Appendix~\ref{sec:sdqa_overview}.
\subsection{Ablation Studies}
In Table~\ref{tab:ablations}, we systematically remove each individual task: Dialogue Listening Comprehension (DLC), Story Listening Comprehension (SLC), and Response Selection (RS). The removal of each auxiliary task results in performance degradation relative to full multi-task tuning, indicating their importance towards improved cross-modal understanding. We observe that removing SLC results in the most performance degradation, which follows the intuition in Figure~\ref{fig:fig1} and \citet{li2018spoken}.

\begin{table}[h]
\centering
\begin{adjustbox}{width=1.0\linewidth,center}
\begin{tabular}{@{}llcc@{}}
\toprule
Approach & Data & Single-Turn Sim. $\uparrow$ & Multi-Turn Sim. $\uparrow$ \\ 
\midrule
\hline
Speech-Qwen MT w/o DLC & 1\% &  53.77 & 53.10 \\
Speech-Qwen MT w/o SLC & 1\% & 52.32 & 51.89 \\
Speech-Qwen MT w/o RS & 1\% & 53.53 & 52.67 \\
Speech-Qwen MT & 1\% & 54.54 & 53.60 \\
\hline
Speech-Qwen MT w/o DLC & 10\% & 65.09 & 64.19 \\
Speech-Qwen MT w/o SLC & 10\% & 64.75 & 64.24 \\
Speech-Qwen MT w/o RS & 10\% &  66.89 & 66.01 \\
Speech-Qwen MT & 10\% & 68.27 & 67.58 \\
\hline
\bottomrule
\end{tabular}
\end{adjustbox}
\vspace{-2.5mm}
\caption{Systematic ablations of each individual task type on ASK-QA in the limited data setting.
\label{tab:ablations}
}
\vspace{-2.5mm}
\end{table}

\vspace{-3mm}
\section{Conclusion}
We propose a data-centric multi-task learning approach which helps improve speech data utilization for MLLM tuning. Tuning on various corpora with Gemini and open-weight MLLMs, we observe consistent performance improvements regardless of model scale and tuning access, surpassing state-of-the-art performance on Spoken-SQuAD with open-weight MLLMs. Future work may build upon our insights by designing new auxiliary tasks, incorporating more expressive TTS approaches (e.g., emotion modeling), or examining action optimization strategies for ASK-QA. Our dataset and synthesis process can also be contributed to post-training data mixtures to improve construction of MLLMs' for improved long-context speech modeling abilities.
\section*{Acknowledgements}
We thank Jinsung Yoon and Ta-Chung Chi for their helpful feedback on our work.
\section*{Limitations}
\textbf{Transcription supervision:} One of the crucial auxiliary tasks in our approach is listening comprehension, as demonstrated by the performance degradation in our ablations (Table~\ref{tab:ablations}). In our implementation, we use ground-truth transcriptions as the target for this generation task. These transcriptions may not be available --- for instance, the ones provided by Spoken-SQuAD and SD-QA were obtained via ASR. Our transcription for ASK-QA is not guaranteed to perfectly match the generated speech either, despite our efforts to filter the data quality (see Appendix~\ref{sec:askqa_details}). It is not clear whether the possibility of slight transcription errors improves model robustness to noise or degrades performance, and this warrants further study in future work.

\textbf{TTS quality:} Our data generation approach is bottlenecked by current capabilities of TTS software. While TTS has greatly improved in recent years in terms of WER, we do still witness generation errors and naturalness issues when working with long context (hence the need for filtering). We are also not at the point in which we have perfect controllability in paralinguistic attributes.

\textbf{Generalization to paralinguistic tasks:} We propose a multi-task approach which can be used to greatly improve performance in SQA. In the three corpora here, listening comprehension proves to be crucial as the primary objective is auditory semantic understanding. However, in more nuanced contexts like task guidance~\cite{schlager2024designing}, it is more important to monitor different paralinguistic aspects of the user such as frustration. 

\textbf{Use in large-scale model post-training:} We believe that our overall data generation process can be useful for improving MLLM post-training. However, verifying this claim is beyond the scope of this work due to computational constraints. We see improved performance on our downstream task after supervised fine-tuning of Gemini, which does indicate positive signal that there are correlations between our training and evaluation data.
\bibliography{acl_latex}

\begin{thebibliography}{77}
\providecommand{\natexlab}[1]{#1}

\bibitem[{Abdin et~al.(2024)Abdin, Aneja, Awadalla, Awadallah, Awan, Bach, Bahree, Bakhtiari, Bao, Behl et~al.}]{abdin2024phi}
Marah Abdin, Jyoti Aneja, Hany Awadalla, Ahmed Awadallah, Ammar~Ahmad Awan, Nguyen Bach, Amit Bahree, Arash Bakhtiari, Jianmin Bao, Harkirat Behl, et~al. 2024.
\newblock Phi-3 technical report: A highly capable language model locally on your phone.
\newblock \emph{arXiv preprint arXiv:2404.14219}.

\bibitem[{Aghajanyan et~al.(2021)Aghajanyan, Gupta, Shrivastava, Chen, Zettlemoyer, and Gupta}]{aghajanyan2021muppet}
Armen Aghajanyan, Anchit Gupta, Akshat Shrivastava, Xilun Chen, Luke Zettlemoyer, and Sonal Gupta. 2021.
\newblock Muppet: Massive multi-task representations with pre-finetuning.
\newblock In \emph{Proceedings of the 2021 Conference on Empirical Methods in Natural Language Processing}, pages 5799--5811.

\bibitem[{Anguera et~al.(2012)Anguera, Bozonnet, Evans, Fredouille, Friedland, and Vinyals}]{anguera2012speaker}
Xavier Anguera, Simon Bozonnet, Nicholas Evans, Corinne Fredouille, Gerald Friedland, and Oriol Vinyals. 2012.
\newblock Speaker diarization: A review of recent research.
\newblock \emph{IEEE Transactions on audio, speech, and language processing}, 20(2):356--370.

\bibitem[{Arora et~al.(2024)Arora, Futami, Jung, Peng, Sharma, Kashiwagi, Tsunoo, Livescu, and Watanabe}]{arora2024universlu}
Siddhant Arora, Hayato Futami, Jee-weon Jung, Yifan Peng, Roshan Sharma, Yosuke Kashiwagi, Emiru Tsunoo, Karen Livescu, and Shinji Watanabe. 2024.
\newblock Universlu: Universal spoken language understanding for diverse tasks with natural language instructions.
\newblock In \emph{Proceedings of the 2024 Conference of the North American Chapter of the Association for Computational Linguistics: Human Language Technologies (Volume 1: Long Papers)}, pages 2754--2774.

\bibitem[{Bhattacharya et~al.(2023)Bhattacharya, Chi, Hirschberg, and Bell}]{bhattacharya2023capturing}
Debasmita Bhattacharya, Jie Chi, Julia Hirschberg, and Peter Bell. 2023.
\newblock Capturing formality in speech across domains and languages.
\newblock \emph{Interspeech 2023}.

\bibitem[{Brown et~al.(2020)Brown, Mann, Ryder, Subbiah, Kaplan, Dhariwal, Neelakantan, Shyam, Sastry, Askell, Agarwal, Herbert-Voss, Krueger, Henighan, Child, Ramesh, Ziegler, Wu, Winter, Hesse, Chen, Sigler, Litwin, Gray, Chess, Clark, Berner, McCandlish, Radford, Sutskever, and Amodei}]{brown2020language}
Tom Brown, Benjamin Mann, Nick Ryder, Melanie Subbiah, Jared~D Kaplan, Prafulla Dhariwal, Arvind Neelakantan, Pranav Shyam, Girish Sastry, Amanda Askell, Sandhini Agarwal, Ariel Herbert-Voss, Gretchen Krueger, Tom Henighan, Rewon Child, Aditya Ramesh, Daniel Ziegler, Jeffrey Wu, Clemens Winter, Chris Hesse, Mark Chen, Eric Sigler, Mateusz Litwin, Scott Gray, Benjamin Chess, Jack Clark, Christopher Berner, Sam McCandlish, Alec Radford, Ilya Sutskever, and Dario Amodei. 2020.
\newblock \href {https://proceedings.neurips.cc/paper_files/paper/2020/file/1457c0d6bfcb4967418bfb8ac142f64a-Paper.pdf} {Language models are few-shot learners}.
\newblock In \emph{Advances in Neural Information Processing Systems}, volume~33, pages 1877--1901. Curran Associates, Inc.

\bibitem[{Caruana(1997)}]{caruana1997multitask}
Rich Caruana. 1997.
\newblock Multitask learning.
\newblock \emph{Machine learning}, 28:41--75.

\bibitem[{Chen et~al.(2022{\natexlab{a}})Chen, Papangelis, Tao, Rosenbaum, Kim, Liu, Yu, and Hakkani-Tur}]{chen2022weakly}
Maximillian Chen, Alexandros Papangelis, Chenyang Tao, Andy Rosenbaum, Seokhwan Kim, Yang Liu, Zhou Yu, and Dilek Hakkani-Tur. 2022{\natexlab{a}}.
\newblock Weakly supervised data augmentation through prompting for dialogue understanding.
\newblock In \emph{NeurIPS 2022 Workshop on Synthetic Data for Empowering ML Research}.

\bibitem[{Chen et~al.(2024)Chen, Sun, Ar{\i}k, and Pfister}]{chen2024learning}
Maximillian Chen, Ruoxi Sun, Sercan~{\"O} Ar{\i}k, and Tomas Pfister. 2024.
\newblock Learning to clarify: Multi-turn conversations with action-based contrastive self-training.
\newblock \emph{arXiv preprint arXiv:2406.00222}.

\bibitem[{Chen et~al.(2023)Chen, Yu, Shi, Awasthi, and Yu}]{chen2023controllable}
Maximillian Chen, Xiao Yu, Weiyan Shi, Urvi Awasthi, and Zhou Yu. 2023.
\newblock Controllable mixed-initiative dialogue generation through prompting.
\newblock In \emph{Proceedings of the 61st Annual Meeting of the Association for Computational Linguistics (Volume 2: Short Papers)}, pages 951--966.

\bibitem[{Chen and Yu(2023)}]{chen2023pre}
Maximillian Chen and Zhou Yu. 2023.
\newblock Pre-finetuning for few-shot emotional speech recognition.
\newblock In \emph{INTERSPEECH}. INTERSPEECH.

\bibitem[{Chen et~al.(2022{\natexlab{b}})Chen, Wang, Chen, Wu, Liu, Chen, Li, Kanda, Yoshioka, Xiao et~al.}]{chen2022wavlm}
Sanyuan Chen, Chengyi Wang, Zhengyang Chen, Yu~Wu, Shujie Liu, Zhuo Chen, Jinyu Li, Naoyuki Kanda, Takuya Yoshioka, Xiong Xiao, et~al. 2022{\natexlab{b}}.
\newblock Wavlm: Large-scale self-supervised pre-training for full stack speech processing.
\newblock \emph{IEEE Journal of Selected Topics in Signal Processing}, 16(6):1505--1518.

\bibitem[{Chu et~al.(2024)Chu, Xu, Yang, Wei, Wei, Guo, Leng, Lv, He, Lin et~al.}]{chu2024qwen2}
Yunfei Chu, Jin Xu, Qian Yang, Haojie Wei, Xipin Wei, Zhifang Guo, Yichong Leng, Yuanjun Lv, Jinzheng He, Junyang Lin, et~al. 2024.
\newblock Qwen2-audio technical report.
\newblock \emph{arXiv preprint arXiv:2407.10759}.

\bibitem[{Chuang et~al.(2020)Chuang, Liu, Lee, and Lee}]{chuang2020speechbert}
Yung-Sung Chuang, Chi-Liang Liu, Hung-yi Lee, and Lin-shan Lee. 2020.
\newblock Speechbert: An audio-and-text jointly learned language model for end-to-end spoken question answering.
\newblock \emph{Interspeech 2020}.

\bibitem[{Deng et~al.(2022)Deng, Lei, Zhang, Lam, and Chua}]{deng2022pacific}
Yang Deng, Wenqiang Lei, Wenxuan Zhang, Wai Lam, and Tat-Seng Chua. 2022.
\newblock Pacific: Towards proactive conversational question answering over tabular and textual data in finance.
\newblock In \emph{Proceedings of the 2022 Conference on Empirical Methods in Natural Language Processing}, pages 6970--6984.

\bibitem[{Deng et~al.(2024)Deng, Liao, Zheng, Yang, and Chua}]{deng2024towards}
Yang Deng, Lizi Liao, Zhonghua Zheng, Grace~Hui Yang, and Tat-Seng Chua. 2024.
\newblock Towards human-centered proactive conversational agents.
\newblock In \emph{Proceedings of the 47th International ACM SIGIR Conference on Research and Development in Information Retrieval}, pages 807--818.

\bibitem[{Devlin(2018)}]{devlin2018bert}
Jacob Devlin. 2018.
\newblock Bert: Pre-training of deep bidirectional transformers for language understanding.
\newblock \emph{arXiv preprint arXiv:1810.04805}.

\bibitem[{Faisal et~al.(2021)Faisal, Keshava, Alam, and Anastasopoulos}]{faisal2021sd}
Fahim Faisal, Sharlina Keshava, Md~Mahfuz~Ibn Alam, and Antonios Anastasopoulos. 2021.
\newblock Sd-qa: Spoken dialectal question answering for the real world.
\newblock In \emph{Findings of the Association for Computational Linguistics: EMNLP 2021}, pages 3296--3315.

\bibitem[{Fatima et~al.(2024)Fatima, Singh, Amipara, and Chaudhary}]{fatima2024accuracy}
Kaleem Fatima, Pinky Singh, Hetal Amipara, and Ganesh Chaudhary. 2024.
\newblock Accuracy of artificial intelligence-based virtual assistants in responding to frequently asked questions related to orthognathic surgery.
\newblock \emph{Journal of Oral and Maxillofacial Surgery}.

\bibitem[{{Gemini Team} et~al.(2023){Gemini Team}, Anil, Borgeaud, Alayrac, Yu, Soricut, Schalkwyk, Dai, Hauth, Millican et~al.}]{team2023gemini}
{Gemini Team}, Rohan Anil, Sebastian Borgeaud, Jean-Baptiste Alayrac, Jiahui Yu, Radu Soricut, Johan Schalkwyk, Andrew~M Dai, Anja Hauth, Katie Millican, et~al. 2023.
\newblock Gemini: a family of highly capable multimodal models.
\newblock \emph{arXiv preprint arXiv:2312.11805}.

\bibitem[{Groeneveld et~al.(2024)Groeneveld, Beltagy, Walsh, Bhagia, Kinney, Tafjord, Jha, Ivison, Magnusson, Wang, Arora, Atkinson, Authur, Chandu, Cohan, Dumas, Elazar, Gu, Hessel, Khot, Merrill, Morrison, Muennighoff, Naik, Nam, Peters, Pyatkin, Ravichander, Schwenk, Shah, Smith, Strubell, Subramani, Wortsman, Dasigi, Lambert, Richardson, Zettlemoyer, Dodge, Lo, Soldaini, Smith, and Hajishirzi}]{OLMo}
Dirk Groeneveld, Iz~Beltagy, Pete Walsh, Akshita Bhagia, Rodney Kinney, Oyvind Tafjord, A.~Jha, Hamish Ivison, Ian Magnusson, Yizhong Wang, Shane Arora, David Atkinson, Russell Authur, Khyathi~Raghavi Chandu, Arman Cohan, Jennifer Dumas, Yanai Elazar, Yuling Gu, Jack Hessel, Tushar Khot, William Merrill, Jacob~Daniel Morrison, Niklas Muennighoff, Aakanksha Naik, Crystal Nam, Matthew~E. Peters, Valentina Pyatkin, Abhilasha Ravichander, Dustin Schwenk, Saurabh Shah, Will Smith, Emma Strubell, Nishant Subramani, Mitchell Wortsman, Pradeep Dasigi, Nathan Lambert, Kyle Richardson, Luke Zettlemoyer, Jesse Dodge, Kyle Lo, Luca Soldaini, Noah~A. Smith, and Hanna Hajishirzi. 2024.
\newblock \href {https://api.semanticscholar.org/CorpusID:267365485} {Olmo: Accelerating the science of language models}.
\newblock \emph{arXiv preprint}.

\bibitem[{Gu et~al.(2021)Gu, Tao, Ling, Xu, Geng, and Jiang}]{gu2021mpc}
Jia-Chen Gu, Chongyang Tao, Zhenhua Ling, Can Xu, Xiubo Geng, and Daxin Jiang. 2021.
\newblock Mpc-bert: A pre-trained language model for multi-party conversation understanding.
\newblock In \emph{Proceedings of the 59th Annual Meeting of the Association for Computational Linguistics and the 11th International Joint Conference on Natural Language Processing (Volume 1: Long Papers)}, pages 3682--3692.

\bibitem[{Guo et~al.(2021)Guo, Zhang, Reddy, and Alikhani}]{guo2021abg}
Meiqi Guo, Mingda Zhang, Siva Reddy, and Malihe Alikhani. 2021.
\newblock Abg-coqa: Clarifying ambiguity in conversational question answering.
\newblock In \emph{3rd Conference on Automated Knowledge Base Construction}.

\bibitem[{Hemphill et~al.(1990)Hemphill, Godfrey, and Doddington}]{hemphill1990atis}
Charles~T Hemphill, John~J Godfrey, and George~R Doddington. 1990.
\newblock The atis spoken language systems pilot corpus.
\newblock In \emph{Speech and Natural Language: Proceedings of a Workshop Held at Hidden Valley, Pennsylvania, June 24-27, 1990}.

\bibitem[{Henderson et~al.(2019)Henderson, Vuli{\'c}, Gerz, Casanueva, Budzianowski, Coope, Spithourakis, Wen, Mrk{\v{s}}i{\'c}, and Su}]{henderson2019training}
Matthew Henderson, Ivan Vuli{\'c}, Daniela Gerz, I{\~n}igo Casanueva, Pawe{\l} Budzianowski, Sam Coope, Georgios Spithourakis, Tsung-Hsien Wen, Nikola Mrk{\v{s}}i{\'c}, and Pei-Hao Su. 2019.
\newblock Training neural response selection for task-oriented dialogue systems.
\newblock In \emph{Proceedings of the 57th Annual Meeting of the Association for Computational Linguistics}, pages 5392--5404.

\bibitem[{Hirschberg(1993)}]{hirschberg1993pitch}
Julia Hirschberg. 1993.
\newblock Pitch accent in context predicting intonational prominence from text.
\newblock \emph{Artificial Intelligence}, 63(1-2):305--340.

\bibitem[{Horvitz(1999)}]{horvitz1999principles}
Eric Horvitz. 1999.
\newblock Principles of mixed-initiative user interfaces.
\newblock In \emph{Proceedings of the SIGCHI conference on Human Factors in Computing Systems}, pages 159--166.

\bibitem[{Huang(2017)}]{huang2017fusionnet}
HY~Huang. 2017.
\newblock Fusionnet: Fusing via fully-aware attention with application to machine comprehension.
\newblock \emph{arXiv preprint arXiv:1711.07341}.

\bibitem[{Jung et~al.(2018)}]{jung2018avoiding}
Jee-weon Jung et~al. 2018.
\newblock Avoiding speaker overfitting in end-to-end dnns using raw waveform for text-independent speaker verification.
\newblock In \emph{Interspeech}.

\bibitem[{Khatri et~al.(2018{\natexlab{a}})Khatri, Hedayatnia, Venkatesh, Nunn, Pan, Liu, Song, Gottardi, Kwatra, Pancholi et~al.}]{khatri2018advancing}
Chandra Khatri, Behnam Hedayatnia, Anu Venkatesh, Jeff Nunn, Yi~Pan, Qing Liu, Han Song, Anna Gottardi, Sanjeev Kwatra, Sanju Pancholi, et~al. 2018{\natexlab{a}}.
\newblock Advancing the state of the art in open domain dialog systems through the alexa prize.
\newblock \emph{arXiv preprint arXiv:1812.10757}.

\bibitem[{Khatri et~al.(2018{\natexlab{b}})Khatri, Venkatesh, Hedayatnia, Gabriel, Ram, and Prasad}]{khatri2018alexa}
Chandra Khatri, Anu Venkatesh, Behnam Hedayatnia, Raefer Gabriel, Ashwin Ram, and Rohit Prasad. 2018{\natexlab{b}}.
\newblock Alexa prize—state of the art in conversational ai.
\newblock \emph{AI magazine}, 39(3):40--55.

\bibitem[{Kim et~al.(2023)Kim, Hessel, Jiang, West, Lu, Yu, Zhou, Le~Bras, Alikhani, Kim et~al.}]{kim2023soda}
Hyunwoo Kim, Jack Hessel, Liwei Jiang, Peter West, Ximing Lu, Youngjae Yu, Pei Zhou, Ronan Le~Bras, Malihe Alikhani, Gunhee Kim, et~al. 2023.
\newblock Soda: Million-scale dialogue distillation with social commonsense contextualization.
\newblock In \emph{The 2023 Conference on Empirical Methods in Natural Language Processing}.

\bibitem[{Kong et~al.(2024)Kong, Goel, Badlani, Ping, Valle, and Catanzaro}]{kongaudio}
Zhifeng Kong, Arushi Goel, Rohan Badlani, Wei Ping, Rafael Valle, and Bryan Catanzaro. 2024.
\newblock Audio flamingo: A novel audio language model with few-shot learning and dialogue abilities.
\newblock In \emph{Forty-first International Conference on Machine Learning}.

\bibitem[{Kurata et~al.(2011)Kurata, Itoh, and Nishimura}]{kurata2011acoustic}
Gakuto Kurata, Nobuyasu Itoh, and Masafumi Nishimura. 2011.
\newblock Acoustic model training with detecting transcription errors in the training data.
\newblock In \emph{INTERSPEECH}, pages 1689--1692.

\bibitem[{Lee et~al.(2019)Lee, Chen, and Lee}]{lee2019mitigating}
Chia-Hsuan Lee, Yun-Nung Chen, and Hung-Yi Lee. 2019.
\newblock Mitigating the impact of speech recognition errors on spoken question answering by adversarial domain adaptation.
\newblock In \emph{ICASSP 2019-2019 IEEE International Conference on Acoustics, Speech and Signal Processing (ICASSP)}, pages 7300--7304. IEEE.

\bibitem[{Li et~al.(2017)Li, Sainath, Narayanan, Caroselli, Bacchiani, Misra, Shafran, Sak, Pundak, Chin et~al.}]{li2017acoustic}
Bo~Li, Tara~N Sainath, Arun Narayanan, Joe Caroselli, Michiel Bacchiani, Ananya Misra, Izhak Shafran, Hasim Sak, Golan Pundak, Kean~K Chin, et~al. 2017.
\newblock Acoustic modeling for google home.
\newblock In \emph{Interspeech}, pages 399--403.

\bibitem[{Li et~al.(2023)Li, Chrysostomou, and Yang}]{li2023speech}
Chen Li, Dimitris Chrysostomou, and Hongji Yang. 2023.
\newblock A speech-enabled virtual assistant for efficient human--robot interaction in industrial environments.
\newblock \emph{Journal of Systems and Software}, 205:111818.

\bibitem[{Li et~al.(2018)Li, Wu, Liu, and Lee}]{li2018spoken}
Chia-Hsuan Li, Szu-Lin Wu, Chi-Liang Liu, and Hung-yi Lee. 2018.
\newblock Spoken squad: A study of mitigating the impact of speech recognition errors on listening comprehension.
\newblock \emph{arXiv preprint arXiv:1804.00320}.

\bibitem[{Li et~al.(2020)Li, Qian, Shi, and Yu}]{li2020end}
Yu~Li, Kun Qian, Weiyan Shi, and Zhou Yu. 2020.
\newblock End-to-end trainable non-collaborative dialog system.
\newblock In \emph{Proceedings of the AAAI Conference on Artificial Intelligence}, volume~34, pages 8293--8302.

\bibitem[{Liu et~al.(2024)Liu, Li, Wu, and Lee}]{liu2024visual}
Haotian Liu, Chunyuan Li, Qingyang Wu, and Yong~Jae Lee. 2024.
\newblock Visual instruction tuning.
\newblock \emph{Advances in neural information processing systems}, 36.

\bibitem[{Ma et~al.(2024)Ma, Yang, Yang, Gao, Wang, Du, Yu, Chen, Zheng, Zhang et~al.}]{ma2024embarrassingly}
Ziyang Ma, Guanrou Yang, Yifan Yang, Zhifu Gao, Jiaming Wang, Zhihao Du, Fan Yu, Qian Chen, Siqi Zheng, Shiliang Zhang, et~al. 2024.
\newblock An embarrassingly simple approach for llm with strong asr capacity.
\newblock \emph{arXiv preprint arXiv:2402.08846}.

\bibitem[{Mehrish et~al.(2023)Mehrish, Majumder, Bharadwaj, Mihalcea, and Poria}]{mehrish2023review}
Ambuj Mehrish, Navonil Majumder, Rishabh Bharadwaj, Rada Mihalcea, and Soujanya Poria. 2023.
\newblock A review of deep learning techniques for speech processing.
\newblock \emph{Information Fusion}, 99:101869.

\bibitem[{Min et~al.(2020)Min, Michael, Hajishirzi, and Zettlemoyer}]{min2020ambigqa}
Sewon Min, Julian Michael, Hannaneh Hajishirzi, and Luke Zettlemoyer. 2020.
\newblock Ambigqa: Answering ambiguous open-domain questions.
\newblock In \emph{Proceedings of the 2020 Conference on Empirical Methods in Natural Language Processing (EMNLP)}, pages 5783--5797.

\bibitem[{Mulholland et~al.(2016)Mulholland, Lopez, Evanini, Loukina, and Qian}]{mulholland2016comparison}
Matthew Mulholland, Melissa Lopez, Keelan Evanini, Anastassia Loukina, and Yao Qian. 2016.
\newblock A comparison of asr and human errors for transcription of non-native spontaneous speech.
\newblock In \emph{2016 IEEE International Conference on Acoustics, Speech and Signal Processing (ICASSP)}, pages 5855--5859. IEEE.

\bibitem[{Nautsch et~al.(2019)Nautsch, Jim{\'e}nez, Treiber, Kolberg, Jasserand, Kindt, Delgado, Todisco, Hmani, Mtibaa et~al.}]{nautsch2019preserving}
Andreas Nautsch, Abelino Jim{\'e}nez, Amos Treiber, Jascha Kolberg, Catherine Jasserand, Els Kindt, H{\'e}ctor Delgado, Massimiliano Todisco, Mohamed~Amine Hmani, Aymen Mtibaa, et~al. 2019.
\newblock Preserving privacy in speaker and speech characterisation.
\newblock \emph{Computer Speech \& Language}, 58:441--480.

\bibitem[{Padmakumar et~al.(2022)Padmakumar, Lausen, Ballesteros, Zha, He, and Karypis}]{padmakumar2022exploring}
Vishakh Padmakumar, Leonard Lausen, Miguel Ballesteros, Sheng Zha, He~He, and George Karypis. 2022.
\newblock Exploring the role of task transferability in large-scale multi-task learning.
\newblock In \emph{Proceedings of the 2022 Conference of the North American Chapter of the Association for Computational Linguistics: Human Language Technologies}, pages 2542--2550.

\bibitem[{Passali et~al.(2022)Passali, Mavropoulos, Tsoumakas, Meditskos, and Vrochidis}]{passali-etal-2022-lard}
Tatiana Passali, Thanassis Mavropoulos, Grigorios Tsoumakas, Georgios Meditskos, and Stefanos Vrochidis. 2022.
\newblock \href {https://aclanthology.org/2022.lrec-1.249} {{LARD}: Large-scale artificial disfluency generation}.
\newblock In \emph{Proceedings of the Thirteenth Language Resources and Evaluation Conference}, pages 2327--2336, Marseille, France. European Language Resources Association.

\bibitem[{Paszke et~al.(2019)Paszke, Gross, Massa, Lerer, Bradbury, Chanan, Killeen, Lin, Gimelshein, Antiga et~al.}]{paszke2019pytorch}
Adam Paszke, Sam Gross, Francisco Massa, Adam Lerer, James Bradbury, Gregory Chanan, Trevor Killeen, Zeming Lin, Natalia Gimelshein, Luca Antiga, et~al. 2019.
\newblock Pytorch: An imperative style, high-performance deep learning library.
\newblock \emph{Advances in neural information processing systems}, 32.

\bibitem[{Pironkov et~al.(2016)}]{pironkov2016speaker}
Gueorgui Pironkov et~al. 2016.
\newblock Speaker-aware long short-term memory multi-task learning for speech recognition.
\newblock In \emph{EUSIPCO}, pages 1911--1915. IEEE.

\bibitem[{Qian et~al.(2018)Qian, Han, Hou, Zhang, Wang, and Li}]{qian2018towards}
Jianwei Qian, Feng Han, Jiahui Hou, Chunhong Zhang, Yu~Wang, and Xiang-Yang Li. 2018.
\newblock Towards privacy-preserving speech data publishing.
\newblock In \emph{IEEE INFOCOM 2018-IEEE Conference on Computer Communications}, pages 1079--1087. IEEE.

\bibitem[{Qian et~al.(2024)Qian, Wan, Tang, Wang, Zhang, Chen, and Yu}]{qian-etal-2024-varbench}
Kun Qian, Shunji Wan, Claudia Tang, Youzhi Wang, Xuanming Zhang, Maximillian Chen, and Zhou Yu. 2024.
\newblock \href {https://doi.org/10.18653/v1/2024.findings-emnlp.946} {{V}ar{B}ench: Robust language model benchmarking through dynamic variable perturbation}.
\newblock In \emph{Findings of the Association for Computational Linguistics: EMNLP 2024}, pages 16131--16161, Miami, Florida, USA. Association for Computational Linguistics.

\bibitem[{Radford et~al.(2023)Radford, Kim, Xu, Brockman, McLeavey, and Sutskever}]{radford2023robust}
Alec Radford, Jong~Wook Kim, Tao Xu, Greg Brockman, Christine McLeavey, and Ilya Sutskever. 2023.
\newblock Robust speech recognition via large-scale weak supervision.
\newblock In \emph{International conference on machine learning}, pages 28492--28518. PMLR.

\bibitem[{Raffel et~al.(2020)Raffel, Shazeer, Roberts, Lee, Narang, Matena, Zhou, Li, and Liu}]{raffel2020exploring}
Colin Raffel, Noam Shazeer, Adam Roberts, Katherine Lee, Sharan Narang, Michael Matena, Yanqi Zhou, Wei Li, and Peter~J Liu. 2020.
\newblock Exploring the limits of transfer learning with a unified text-to-text transformer.
\newblock \emph{Journal of machine learning research}, 21(140):1--67.

\bibitem[{Rajpurkar et~al.(2016)Rajpurkar, Zhang, Lopyrev, and Liang}]{rajpurkar2016squad}
Pranav Rajpurkar, Jian Zhang, Konstantin Lopyrev, and Percy Liang. 2016.
\newblock Squad: 100,000+ questions for machine comprehension of text.
\newblock In \emph{Proceedings of the 2016 Conference on Empirical Methods in Natural Language Processing}, pages 2383--2392.

\bibitem[{Rasley et~al.(2020)Rasley, Rajbhandari, Ruwase, and He}]{rasley2020deepspeed}
Jeff Rasley, Samyam Rajbhandari, Olatunji Ruwase, and Yuxiong He. 2020.
\newblock Deepspeed: System optimizations enable training deep learning models with over 100 billion parameters.
\newblock In \emph{Proceedings of the 26th ACM SIGKDD International Conference on Knowledge Discovery \& Data Mining}, pages 3505--3506.

\bibitem[{Reimers(2019)}]{reimers2019sentence}
N~Reimers. 2019.
\newblock Sentence-bert: Sentence embeddings using siamese bert-networks.
\newblock \emph{arXiv preprint arXiv:1908.10084}.

\bibitem[{Risch et~al.(2021)Risch, M{\"o}ller, Gutsch, and Pietsch}]{risch-etal-2021-semantic}
Julian Risch, Timo M{\"o}ller, Julian Gutsch, and Malte Pietsch. 2021.
\newblock \href {https://doi.org/10.18653/v1/2021.mrqa-1.15} {Semantic answer similarity for evaluating question answering models}.
\newblock In \emph{Proceedings of the 3rd Workshop on Machine Reading for Question Answering}, pages 149--157, Punta Cana, Dominican Republic. Association for Computational Linguistics.

\bibitem[{Roberts et~al.(2023)Roberts, Thakur, Herlihy, White, and Dooley}]{roberts2023cutoff}
Manley Roberts, Himanshu Thakur, Christine Herlihy, Colin White, and Samuel Dooley. 2023.
\newblock To the cutoff... and beyond? a longitudinal perspective on llm data contamination.
\newblock In \emph{The Twelfth International Conference on Learning Representations}.

\bibitem[{Schlager and Feiner(2024)}]{schlager2024designing}
Bettina Schlager and Steven~K Feiner. 2024.
\newblock Designing non-humanoid virtual assistants for task-oriented ar environments.
\newblock In \emph{2024 IEEE Conference on Virtual Reality and 3D User Interfaces Abstracts and Workshops (VRW)}, pages 1017--1018. IEEE.

\bibitem[{Seedat et~al.(2022)Seedat, Imrie, and van~der Schaar}]{seedat2022dc}
Nabeel Seedat, Fergus Imrie, and Mihaela van~der Schaar. 2022.
\newblock Dc-check: A data-centric ai checklist to guide the development of reliable machine learning systems.
\newblock \emph{arXiv preprint arXiv:2211.05764}.

\bibitem[{Sharma et~al.(2024)Sharma, Keh, Mitchell, Finn, Arora, and Kollar}]{sharma2024critical}
Archit Sharma, Sedrick Keh, Eric Mitchell, Chelsea Finn, Kushal Arora, and Thomas Kollar. 2024.
\newblock A critical evaluation of ai feedback for aligning large language models.
\newblock \emph{arXiv preprint arXiv:2402.12366}.

\bibitem[{Shih et~al.(2024)Shih, Chung, Pai, Hsu, Lin, Li, and Lee}]{shih2023gsqa}
Min-Han Shih, Ho-Lam Chung, Yu-Chi Pai, Ming-Hao Hsu, Guan-Ting Lin, Shang-Wen Li, and Hung-yi Lee. 2024.
\newblock Gsqa: An end-to-end model for generative spoken question answering.
\newblock \emph{Interspeech 2024}.

\bibitem[{Su and Fung(2020)}]{su2020improving}
Dan Su and Pascale Fung. 2020.
\newblock \href {https://doi.org/10.1109/ICASSP40776.2020.9053979} {Improving spoken question answering using contextualized word representation}.
\newblock In \emph{ICASSP 2020 - 2020 IEEE International Conference on Acoustics, Speech and Signal Processing (ICASSP)}, pages 8004--8008.

\bibitem[{Talmor et~al.(2019)Talmor, Herzig, Lourie, and Berant}]{talmor-etal-2019-commonsenseqa}
Alon Talmor, Jonathan Herzig, Nicholas Lourie, and Jonathan Berant. 2019.
\newblock \href {https://doi.org/10.18653/v1/N19-1421} {{C}ommonsense{QA}: A question answering challenge targeting commonsense knowledge}.
\newblock In \emph{Proceedings of the 2019 Conference of the North {A}merican Chapter of the Association for Computational Linguistics: Human Language Technologies, Volume 1 (Long and Short Papers)}, pages 4149--4158, Minneapolis, Minnesota. Association for Computational Linguistics.

\bibitem[{Von~Ahn(2013)}]{von2013duolingo}
Luis Von~Ahn. 2013.
\newblock Duolingo: learn a language for free while helping to translate the web.
\newblock In \emph{Proceedings of the 2013 international conference on Intelligent user interfaces}, pages 1--2.

\bibitem[{Wang et~al.(2020)}]{wang2020spoken}
Tao Wang et~al. 2020.
\newblock Spoken content and voice factorization for few-shot speaker adaptation.
\newblock In \emph{INTERSPEECH}, pages 796--800.

\bibitem[{Wolf et~al.(2020)Wolf, Debut, Sanh, Chaumond, Delangue, Moi, Cistac, Rault, Louf, Funtowicz et~al.}]{wolf2020transformers}
Thomas Wolf, Lysandre Debut, Victor Sanh, Julien Chaumond, Clement Delangue, Anthony Moi, Pierric Cistac, Tim Rault, R{\'e}mi Louf, Morgan Funtowicz, et~al. 2020.
\newblock Transformers: State-of-the-art natural language processing.
\newblock In \emph{Proceedings of the 2020 conference on empirical methods in natural language processing: system demonstrations}, pages 38--45.

\bibitem[{Wu et~al.(2023{\natexlab{a}})Wu, Gung, Shu, and Zhang}]{wu2023diacttod}
Qingyang Wu, James Gung, Raphael Shu, and Yi~Zhang. 2023{\natexlab{a}}.
\newblock Diacttod: Learning generalizable latent dialogue acts for controllable task-oriented dialogue systems.
\newblock In \emph{Proceedings of the 24th Annual Meeting of the Special Interest Group on Discourse and Dialogue}, pages 255--267.

\bibitem[{Wu et~al.(2024)Wu, Fei, Qu, Ji, and Chua}]{wunext}
Shengqiong Wu, Hao Fei, Leigang Qu, Wei Ji, and Tat-Seng Chua. 2024.
\newblock Next-gpt: Any-to-any multimodal llm.
\newblock In \emph{Forty-first International Conference on Machine Learning}.

\bibitem[{Wu et~al.(2023{\natexlab{b}})Wu, Parish, Cheng, Min, Ammanabrolu, Ostendorf, and Hajishirzi}]{wu2023inscit}
Zeqiu Wu, Ryu Parish, Hao Cheng, Sewon Min, Prithviraj Ammanabrolu, Mari Ostendorf, and Hannaneh Hajishirzi. 2023{\natexlab{b}}.
\newblock Inscit: Information-seeking conversations with mixed-initiative interactions.
\newblock \emph{Transactions of the Association for Computational Linguistics}, 11:453--468.

\bibitem[{Yang et~al.(2024{\natexlab{a}})Yang, Yang, Hui, Zheng, Yu, Zhou, Li, Li, Liu, Huang et~al.}]{yang2024qwen2}
An~Yang, Baosong Yang, Binyuan Hui, Bo~Zheng, Bowen Yu, Chang Zhou, Chengpeng Li, Chengyuan Li, Dayiheng Liu, Fei Huang, et~al. 2024{\natexlab{a}}.
\newblock Qwen2 technical report.
\newblock \emph{arXiv preprint arXiv:2407.10671}.

\bibitem[{Yang et~al.(2024{\natexlab{b}})Yang, Shi, Le, Hsu, and Tjandra}]{yang2024audiobox}
Mu~Yang, Bowen Shi, Matthew Le, Wei-Ning Hsu, and Andros Tjandra. 2024{\natexlab{b}}.
\newblock Audiobox tta-rag: Improving zero-shot and few-shot text-to-audio with retrieval-augmented generation.
\newblock \emph{arXiv preprint arXiv:2411.05141}.

\bibitem[{You et~al.(2022)You, Chen, Liu, Ge, Wu, and Zou}]{you-etal-2022-end}
Chenyu You, Nuo Chen, Fenglin Liu, Shen Ge, Xian Wu, and Yuexian Zou. 2022.
\newblock \href {https://doi.org/10.18653/v1/2022.findings-naacl.91} {End-to-end spoken conversational question answering: Task, dataset and model}.
\newblock In \emph{Findings of the Association for Computational Linguistics: NAACL 2022}, pages 1219--1232, Seattle, United States. Association for Computational Linguistics.

\bibitem[{Yu et~al.(2022)Yu, Fu, and Zhang}]{yu2022speaker}
Nan Yu, Guohong Fu, and Min Zhang. 2022.
\newblock Speaker-aware discourse parsing on multi-party dialogues.
\newblock In \emph{Proceedings of the 29th International Conference on Computational Linguistics}, pages 5372--5382.

\bibitem[{Yu et~al.(2023)Yu, Chen, and Yu}]{yu2023prompt}
Xiao Yu, Maximillian Chen, and Zhou Yu. 2023.
\newblock Prompt-based monte-carlo tree search for goal-oriented dialogue policy planning.
\newblock In \emph{Proceedings of the 2023 Conference on Empirical Methods in Natural Language Processing}, pages 7101--7125.

\bibitem[{Yu et~al.(2024)Yu, Wu, Li, and Yu}]{yu2024lions}
Xiao Yu, Qingyang Wu, Yu~Li, and Zhou Yu. 2024.
\newblock Lions: An empirically optimized approach to align language models.
\newblock In \emph{Proceedings of the 2024 Conference on Empirical Methods in Natural Language Processing}, pages 8732--8753.

\bibitem[{Zheng et~al.(2024)Zheng, Chiang, Sheng, Li, Zhuang, Wu, Zhuang, Li, Lin, Xing et~al.}]{zhenglmsys}
Lianmin Zheng, Wei-Lin Chiang, Ying Sheng, Tianle Li, Siyuan Zhuang, Zhanghao Wu, Yonghao Zhuang, Zhuohan Li, Zi~Lin, Eric Xing, et~al. 2024.
\newblock Lmsys-chat-1m: A large-scale real-world llm conversation dataset.
\newblock In \emph{The Twelfth International Conference on Learning Representations}.

\end{thebibliography}
\clearpage
\appendix
\renewcommand{\thetable}{A\arabic{table}}
\renewcommand{\thefigure}{A\arabic{figure}}
\renewcommand{\theequation}{A\arabic{equation}}
\section{Additional Related Work}
\paragraph{Spoken question answering} is a fundamental skill for intelligent spoken conversational agents~\cite{khatri2018advancing, zhenglmsys}. Many tasks have been proposed in order to measure models' ability to understand spoken context~\cite{li2018spoken,shih2023gsqa} and spoken requests~\cite{faisal2021sd}. 
Previously, most approaches to SQA focused on span prediction using ``cascade'' approaches which include an intermediate step invoking an Automatic Speech Recognition (ASR) module followed by a fine-tuned a text classification model~\citet{chuang2020speechbert,li2018spoken,su2020improving} like BERT~\cite{devlin2018bert}. It is increasingly desirable to develop end-to-end pipelines to solve SQA tasks~\cite{shih2023gsqa}, particularly with the rise of generalist MLLMs~\cite{wunext}. Such end-to-end models are desirable in speech as they afford opportunities to directly encode useful information in acoustic signals such as speaking rate, pitch, or emotions. In our work, we focus on improving methods for end-to-end SQA using both closed-weight and open-weight MLLMs.

\paragraph{Mixed-initiative conversations}
require each interlocutor to control the interaction flow~\cite{horvitz1999principles} through the use of various pragmatic actions~\cite{chen2023controllable} such as clarifying questions, which can lead to to better goal completion outcomes~\cite{guo2021abg,min2020ambigqa,wu2023inscit}. Many works focus on planning these explicit pragmatic actions~\cite{deng2024towards,yu2023prompt}, whereas other works focus on implicit~~\cite{chen2024learning} and continuous space actions~\cite{wu2023diacttod}, and end-to-end generation capabilities in such settings~\cite{li2020end, deng2022pacific}. While there have been recent efforts in designing multi-turn SQA corpora~\cite{you-etal-2022-end}, to our knowledge, there is not yet any mixed-initiative conversation environment for speech, despite there being many additional acoustic features which may introduce ambiguity ~\cite{kurata2011acoustic,mulholland2016comparison}. In our work, we develop the first-ever conversational SQA corpus which requires the ability to disambiguate requests and reason about clarification questions.

\paragraph{Adapting models with limited speech data} has received much attention due to well-known problem of speaker overfitting across a variety of tasks ranging from grammatical error correction~\cite{wang2020spoken} to speaker verification~\cite{jung2018avoiding}. This problem is frequently addressed with the assistance of multi-task learning~\cite{caruana1997multitask}. \citet{pironkov2016speaker} proposed a multi-task objective in which they simultaneously train a network for both ASR (their downstream task) and speaker classification.
\citet{chen2023pre} found large downstream task performance improvements on speech classification tasks following a stage of multi-task pre-finetuning. In our work, we view multi-task learning through a data-centric lens. While multi-task pre-finetuning relies on additional datasets~\cite{aghajanyan2021muppet,padmakumar2022exploring}, our approach improves the utilization of a fixed set of speech recordings by introducing auxiliary tasks designed to improve the cross-modal understanding capabilities of MLLMs.

\begin{figure*}[t]
    \centering
    \includegraphics[width=\linewidth]{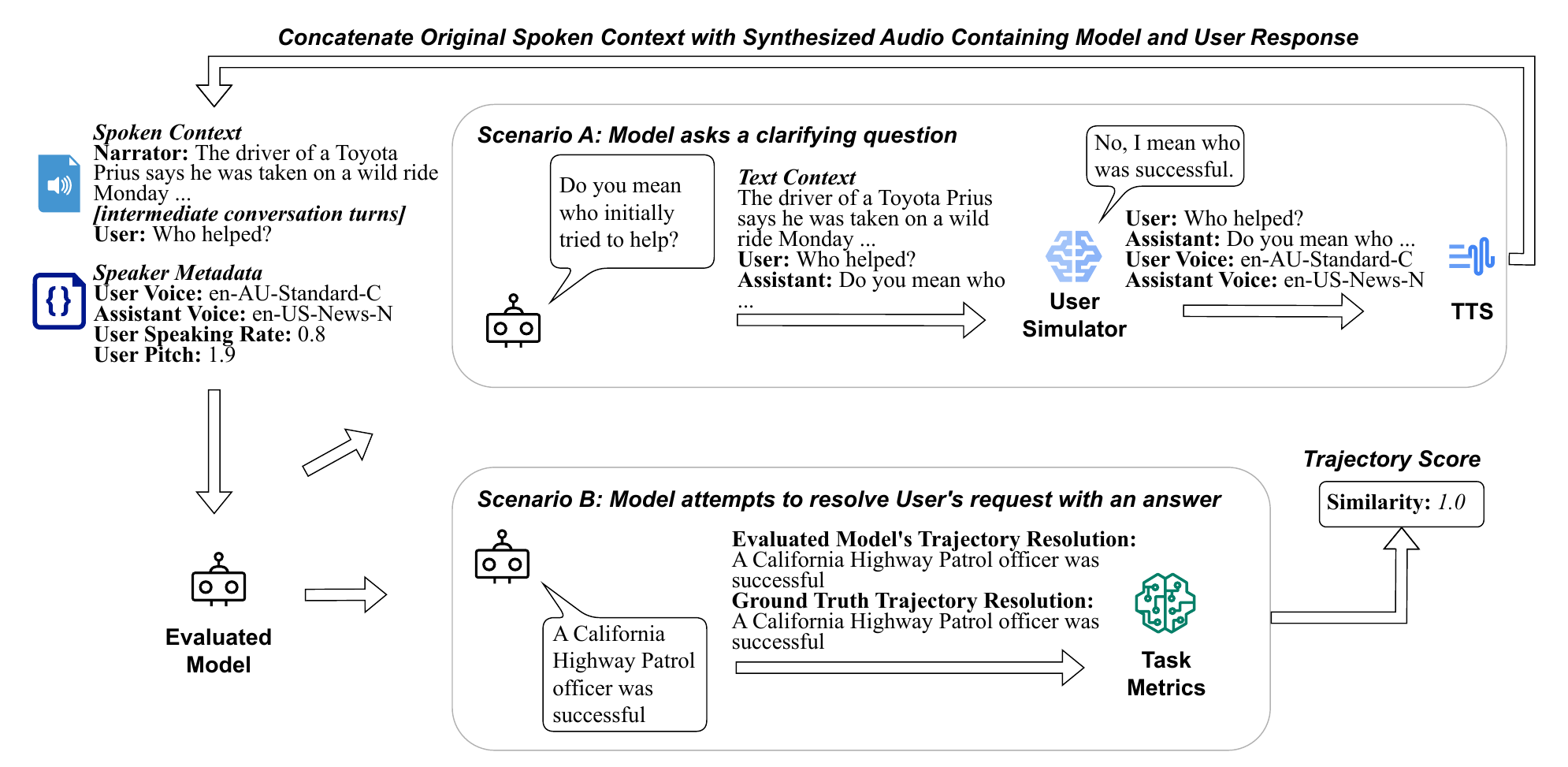}
    \caption{\textbf{Multi-turn evaluation pipeline for ASK-QA.} A model is given an audio recording containing the spoken story and spoken conversation. It is tasked with providing the correct response. While the model response is a clarifying question (as determined by a prompted Action Classifier), the model-generated response is appended to a textual version of the conversation history and shown to a user simulator. The user simulator provides a coherent response to the clarifying question, and these two generated turns are synthesized using TTS to create a new spoken context. This process repeats until the model response is not a clarifying question.}
    \label{fig:eval-pipeline}
\end{figure*}
\section{Additional Details on ASK-QA}
\label{sec:askqa_details}
Here, we provide several additional details on ASK-QA. First we describe the construction of the dataset in detail. Then, we demonstrate the quality of our corpus quantitatively as well as qualitatively. Specifically, we find that the audio quality is very high with strong faithfulness to the original transcript.

\subsection{Dataset Construction}
\paragraph{Overview} As mentioned in Section~\ref{sec:experiments}, ASK-QA is constructed using Abg-CoQA as a starting point~\cite{guo2021abg}. Abg-CoQA is a textual conversational QA task, but as it is span-based, it does not provide very natural dialogue. Each instance consists of a passage which serves as some necessary contextual knowledge, and each conversation consists of dialogue turns where a user asks questions and an assistant is supposed to provide the correct answer or ask a clarifying question if the user's question is contextually ambiguous. The first step we take is to paraphrase each question using Gemini 1.5 Pro to convert the task into free-form QA generation.

\paragraph{Setting speaker roles} Each written conversation can be considered a machine reading comprehension task. We break them down into three components: a story, the set of user questions, and the set of corresponding assistant responses. Our goal is to convert this into a listening comprehension task with two speakers having a conversation about some spoken context. Thus, for each conversation, we construct three unique speaker profiles to represent a story narrator, a user, and an assistant. 

\paragraph{Speaker profiles} Earlier works~\cite{li2018spoken,you-etal-2022-end} used commercial text-to-speech (TTS) software to synthesize speech, but at the time there were relatively high word error rate (WER) with limited options for customization. As a result, such corpora only feature a single synthetic voice without varied acoustic features (e.g. speaking rate, pitch). Here, we construct a much more diverse corpus using modern TTS solution from Google Cloud\footnote{https://cloud.google.com/text-to-speech}. To create user speaker profiles, we aim to maximize diversity and thus sample from 38 unique voice types spanning four different accents from English-speaking countries (US, AU, GB, IN). We also randomly sample user speaking rates and pitches from a truncated normal distribution. The mean of each is set to the default value of the API endpoint. For the assistant and narrator speaker profiles, we aim to create professional-sounding dialogue, instead sampling from 26 different ``news'' and ``studio'' voices. 

\paragraph{Text-to-Speech pipeline} As per Figure~\ref{fig:data_generation}, we then apply TTS to synthesize the story and each dialogue turn sequentially, using the appropriate speaker profile. We then concatenate the resulting audio files into a single spoken conversation. We do not adjust the default speaking rates and pitches. Following the suggestions of earlier work in text-based data synthesis~\cite{chen2022weakly,kim2023soda}, we apply weakly supervised filtering to ensure that the synthesized speech is high quality. If any synthesized speech exceeds a WER of 0.20 (as determined by Whisper-Large v3; \citet{radford2023robust}), we retry the synthesis process. If it fails three times, we discard the conversation sample. We finally randomly insert white noise into the audio by drawing from a Gaussian distribution (with an average signal to noise ratio of 21.75dB). The result is a unique speech CQA dataset with disfluencies, multiple speakers, and long audio context. The contributions of ASK-QA compared to other existing SQA datasets are in Table~\ref{tab:data_comparison}.

\subsection{Additional Evaluation Details}
\label{sec:evaluation_details}
\paragraph{Single-turn evaluation:} We follow the standard single-turn evaluation setting with pre-determined inputs similar to existing conversational QA tasks \citet{guo2021abg,deng2022pacific}. For an evaluation instance, an agent must produce a correct response conditioned on the speech recording. In ASK-QA, the speech recording contains both the knowledge context and a multi-turn conversation context. We compare the generated answer against the ground truth response.

\paragraph{Multi-turn evaluation:}
\citet{chen2024learning} propose an automatic multi-turn evaluation for Abg-CoQA, in which an agent dynamically interacts with a user simulator to work towards the goal of a conversation. Inspired by this idea, we design the first ever dynamic \textit{speech} evaluation, which is summarized in Figure~\ref{fig:eval-pipeline}. 

The evaluation starts by providing the evaluated MLLM with the audio file (see ``Spoken Context'' in Figure~\ref{fig:eval-pipeline}) and the text instruction. Next, an action classifier determines whether the agent response is a clarifying question or a direct answer. The action classifier is an LLM which is directly prompted with the textual conversation context, the candidate response, and the action candidates (CLARIFY/ANSWER). In this work, we use Gemini 1.5 Pro. If it is a clarifying question, we simulate a user-side response to the clarifying question using a user simulator. This user simulator is a high-capacity LLM (here, Gemini 1.5 Pro as well) provided with the textual conversation context, and a summarization of the user's ground truth information-seeking intent. Our implementation of these is the same as for Abg-CoQA in \citet{chen2024learning}.

In the text setting, they concatenate the original context with the newly simulated trajectory and provides this as a new prompt to the evaluated agent. Here, we apply the same TTS process as in Section~\ref{sec:speech_generation} using the same speaker profiles for this conversation to dynamically create another fully spoken input. We then sample another response from the evaluated agent until a direct answer attempt is provided, or the maximum rollout depth is reached (here, we set this to three clarification turns). We apply the evaluation metric to the final response.
\begin{table}[!t]
\small
\begin{tabular}{p{0.1\linewidth}p{0.05\linewidth}p{0.15\linewidth}p{0.45\linewidth}}
Version & Turn No. & Speaker & Content \\ \midrule
Abg-CoQA & 1 & User & what was the others reaction? \\
& 2 & Assistant & Everyone was happy \\
& 3 & User & was he hungry? \\
& 4 & Assistant & Do you mean at first? \\
& 5a & User & Yes \\
& 6a & Assistant & Yes \\
& 5b & User & No, I mean ultimately. \\
& 6b & Assistant & No, he was not hungry anymore. \\
\midrule
ASK-QA & 1 & User & what was the others reaction? \\
& 2 & Assistant & Everyone was happy. \\
& 3 & User & was he hungry? \\
& 4 & Assistant & Do you mean at first? \\
& 5a & User & Yes \\
& 6a & Assistant & \textbf{Yes, he was hungry at first.} \\
& 5b & User & No, I mean ultimately. \\
& 6b & Assistant & No, he was not hungry anymore. \\
\bottomrule
\end{tabular}
\caption{\textbf{Comparison of the dialogue turns from a conversation in Abg-CoQA and the paraphrased version in ASK-QA.} Bold: text paraphrased using Gemini. We do not perturb the user turns in order to maintain the original linguistic diversity. However, for the Assistant turns, we paraphrase the language concisely if necessary in order to ensure that the speech is grammatically correct. Here, the only dialogue turn that differs is 6a. (a) and (b) denote differing trajectories, which are the turns that the Assistant has to navigate successfully during evaluation.}
\label{tab:abgcoqa_comparison1}
\end{table}

\begin{table}[!t]
\small
\begin{tabular}{p{0.1\linewidth}p{0.05\linewidth}p{0.15\linewidth}p{0.45\linewidth}}
Version & Turn No. & Speaker & Content \\ \midrule
Abg-CoQA & 1 & User & Are they related? \\
& 2 & Assistant & yes \\
& 3 & User & How? \\
& 4 & Assistant & brothers \\
& 5 & User & Where do they put the lemonade stand? \\
& 6 & Assistant & by the sidewalk \\
\midrule
ASK-QA & 1 & User & Are they related? \\
& 2 & Assistant & Yes, they are related. \\
& 3 & User & How? \\
& 4 & Assistant & \textbf{They are brothers.} \\
& 5 & User & Where do they put the \textit{the} lemonade stand ? \\
& 6 & Assistant & \textbf{They put the lemonade stand by the sidewalk.} \\
\bottomrule
\end{tabular}
\caption{\textbf{A modified conversation in ASK-QA.} Bold: paraphrased text using Gemini. Italics: repeat disfluency injected using LARD~\cite{passali-etal-2022-lard}. }
\label{tab:askqa_disfluency}
\end{table}
\subsection{Conversation Examples}
In Table~\ref{tab:abgcoqa_comparison1}, we provide a simple example of how a conversation in ASK-QA is paraphrased from the original conversation in Abg-CoQA (with the passage context omitted). One turn is grammaically incorrect and so it is paraphrased using Gemini 1.5 Pro.

In Table~\ref{tab:askqa_disfluency}, we provide an example of a conversation in ASK-QA with more perturbations from Abg-CoQA. Turn 4 is rephrased as a complete sentence. Turn 5 injects a repeat disfluency into the User-side speech. Turn 6 is also rephrased as a complete sentence.

Table~\ref{askqa_rg} provides a full example of a full example from the ASK-QA dataset. We include the passage context, as well as the provided dialogue excerpt. We denote the input modalities as well as our instruction for response generation using the MLLM.

Our supplementary material contains an example of an evaluation instance in our dataset.
\begin{figure*}[!ht]
    \centering
    \includegraphics[width=\textwidth]{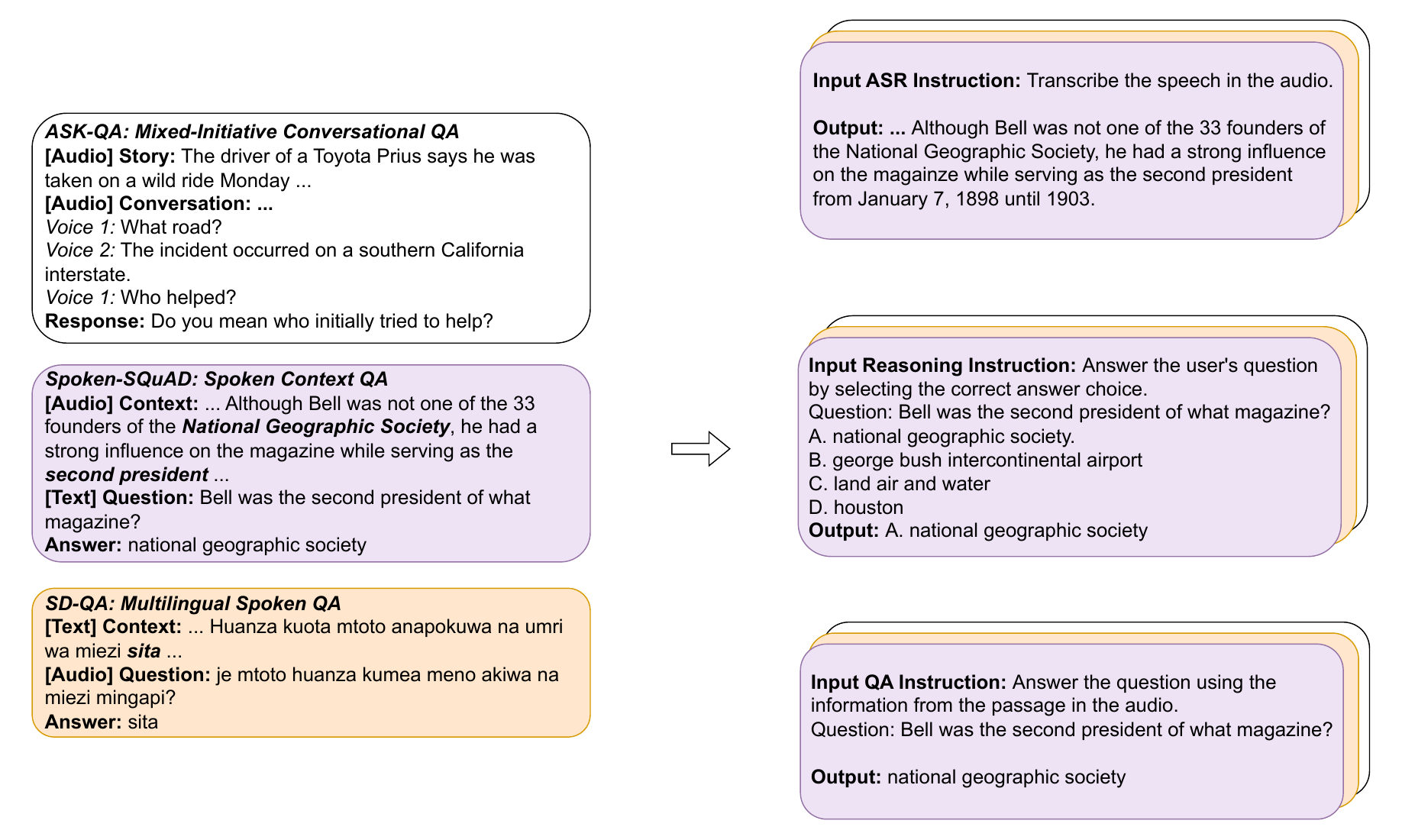}
    \caption{\textbf{Creating multi-task data from individual SQA training instances.} Left: examples of instance metadata from the three SQA datasets used in this paper. Right: for each speech-QA pairing, we are able to form three tasks designed to teach MLLMs' cross-modal reasoning ability.}
    \label{fig:multitask_mapping}
\end{figure*}
\section{Efficient Multimodal Adapters via Audio Representation Projection}
\label{sec:adapter_details}
As demonstrated in Section~\ref{sec:experiments}, our data-centric approach is easily applicable to both settings with access to tuning APIs for closed-source MLLMs like Gemini, and settings with access to open-weight models for each modality. Here, we describe our approach in the open-weight scenario.

Textual instructions serve as a highly controllable interface, and as such, recent work has found much success in unifying multiple modalities with large pre-trained decoder-only language models~\cite{liu2024visual, arora2024universlu, kongaudio}. These works aim to leverage the impressive instruction-following capabilities of LLMs to interpret additional modalities (e.g. vision, speech, video etc.) by effectively mapping their representations to LLM input space.

\paragraph{Architecture:}
In our work, we consider the high-level architecture presented in \citet{ma2024embarrassingly}. We projecting the speech input represented by an audio encoder into the embedding space of an LLM to improve performance on ASR tasks, only tuning the weights of a linear projection layer and freezing the other model components.\footnote{As in \citet{ma2024embarrassingly}, the projection layer consists of merely 17.8M parameters for the proposed models.} As described in Section~\ref{sec:experiments}, our speech encoder is WavLM-Large~\cite{chen2022wavlm}. We primarily experimented with tuning Qwen 2.5-Instruct~\cite{yang2024qwen2} with 7B parameters as our base decoder-only LLM. We also experimented with Phi 3.5 Mini~\cite{abdin2024phi} with 3B parameters in Table~\ref{tab:squad_results}. These MLLMs are referred to as Speech-Qwen and Speech-Phi, respectively. We tune this adapter using standard cross-entropy loss. Details on our tuning experiments are provided in Appendix~\ref{sec:adapter_details}.

\paragraph{Projection Layer Pre-training:}
While this projection layer is tuned directly on the target ASR task in \citet{ma2024embarrassingly}, we find that this approach may struggle with direct single-task fine-tuning on our more difficult SQA tasks which do not have the same abundance of data. Similarly to how visual MLLMs are often pre-trained on image captioning~\cite{liu2024visual}, we pretrain the projection layer for one epoch on large-scale ASR data.

\section{Additional Details on Data-Centric Multi-Task Learning}
Figure~\ref{fig:multitask_mapping} provides a high-level overview of our multi-task learning approach. On the left, we show examples of each of our SQA corpora used for experimentation. At a high level, each corpus consists of passage and a conversation. In ASK-QA, the contextual inputs are fully spoken. In Spoken-SQuAD, the knowledge is spoken while the question is written. In SD-QA, the knowledge is written while the question is spoken (in multiple languages and regional dialects). 

Regardless of the input modalities, each instance can be mapped to new data instances representing the auxiliary tasks in Section~\ref{sec:recipe}. The visible examples on the right side of Figure~\ref{fig:multitask_mapping} are our multi-task instances for Spoken-SQuAD. The top-right panel is our Listening Comprehension task, our middle-right panel is our Cross-Modal Commonsense Reasoning task, and our bottom right task is the standard QA task (which is just reorganized from the middle-left panel).

\subsection{Multi-Task Training Examples for ASK-QA}
\label{sec:multitask_implementation}
We provide concrete examples of the auxiliary tasks for a single instance of ASK-QA in Tables~\ref{askqa_rs},~\ref{askqa_dlc},~\ref{askqa_slc},~\ref{askqa_rg}. Each of these tables has the exact same speech recording. Table~\ref{askqa_rs} demonstrates how the ground-truth answer is joined to negatively sampled answers from other QA pairings to form the response selection task. Table~\ref{askqa_rg} is similar and demonstrates the textual instruction used to steer the MLLM to directly generate the ground-truth answer. As stated in Section~\ref{sec:askqa_experiments}, we break the Listening Comprehension task into two components since each recording comprises a narrated story and a conversation. Table~\ref{askqa_dlc} demonstrates steering the MLLM to transcribe the conversation. Table~\ref{askqa_slc} demonstrates steering the MLLM to transcribe the narrated story.

\begin{table}[t]
\small
\centering
\begin{adjustbox}{width=1.0\linewidth,center}
\begin{tabular}{@{}lllcc@{}}
\toprule
Base Model & App. & Data & EM $\uparrow$ & F1 $\uparrow$ \\ \midrule
Gemini Pro & -- & 0\% & 42.44 & 64.18\\
Gemini Pro & ST & 1\% & 46.33 & 67.74 \\
Gemini Pro & MT & 1\% & \textbf{55.24} & \textbf{70.73} \\
\hline
Gemini Pro & ST & 10\% & 62.79 & 77.80\\
Gemini Pro & MT & 10\% & \textbf{63.04} & \textbf{79.02} \\
\hline
Gemini Pro & ST & 100\% & 63.10 & 78.15\\
Gemini Pro & MT & 100\% & \textbf{64.17} & \textbf{79.06} \\
\hline
\midrule
Speech-Qwen & ST & 1\% & 13.44 & 25.28 \\
Speech-Qwen & MT & 1\% & \textbf{24.70} & \textbf{38.63}\\
\hline
Speech-Qwen & ST & 10\% & 29.70 & 43.84 \\
Speech-Qwen & MT & 10\% & \textbf{39.92} & \textbf{54.35} \\
\hline
Speech-Qwen & ST & 100\% & 46.83 & 61.76\\
Speech-Qwen & MT & 100\% & \textbf{49.54} & \textbf{64.94 }\\
\bottomrule
\end{tabular}
\end{adjustbox}
\caption{\textbf{Experimental results comparing single-task SFT and our proposed multi-task approach on SD-QA's test set}.
\label{tab:sdqa_results}
}
\end{table}
\section{Extended Experimental Results}
\label{sec:extended_experiments}
Due to space constraints in the main text, we describe our additional experiments in this section. In particular, we examine a data-centric multi-task learning on a third corpus which features multi-lingual SQA scenarios, and we look at additional baseline comparisons and base MLLMs.
\subsection{SD-QA: Textual Knowledge and Spoken Questions}
\label{sec:sdqa_overview}
We examine the setting where the single-turn QA context is provided in the recorded speech, and the knowledge necessary to answer the question correctly is provided in the text.
\subsubsection{Dataset}
SD-QA~\citep{faisal2021sd} is a large single-turn SQA benchmark with diverse data -- spanning 5 languages (Arabic, Bengali, English, Kiswahili, and Korean) and 24 regional dialects. 
SD-QA is also proposed as a span-based QA task, but we apply our end-to-end generative approach as in Section~\ref{sec:experiments}. We tune our models on up to 9,008 of the 10,0008 samples made available for training, withholding the remaining samples for validation. We evaluate our approach on the 12,975 evaluation samples.

\paragraph{Findings:} We evaluate performance on SD-QA in terms of exact match and token-based F1. Consistent with our findings in Section~\ref{sec:experiments}, we see that our multi-task approach is able to outperform single-task tuning in all evaluation settings. This is inclusive of experiments with Gemini Pro as the base MLLM for tuning. We see a large 16.13\% relative improvement (46.33 to 55.24) for exact match in the limited data setting with Gemini. 

We observe that Gemini Pro is already a strong base MLLM, achieving competitive zero-shot performance on this corpus. This is likely due to it having a strong initialization on multilingual ASR. We see that our Speech-Qwen model is able to outperform zero-shot Gemini using our multi-task approach with full data. We also observe up to a 52.8\% relative improvement over single-task tuning with Speech-Qwen in the 1\% data regime. This is consistent with findings from \citet{chen2023pre} in which pre-finetuning yields strong improvements in the extremely limited data regime. Overall, the particularly large performance improvements on this particular corpus may be an indication that the base models have not been trained on as much multi-lingual data.

\subsection{Additional Experiments on ASK-QA}
Our main findings and results are presented in Section~\ref{sec:askqa_experiments}. Here, we present our full results on ASK-QA. Specifically, we additionally examine the efficacy of our approach with an additional closed-source MLLM, Gemini Flash. Our results in Figure~\ref{fig:askqa_multiturn_results} also highlight  trajectory-level similarity, and here, we also present results on single-turn evaluation.
\begin{table}[t]
\centering
\begin{adjustbox}{width=1.0\linewidth,center}
\begin{tabular}{@{}lllcc@{}}
\toprule
Base Model & App. & Data & Single-Turn Sim. $\uparrow$ & Multi-Turn Sim. $\uparrow$ \\ \midrule
\hline
Gemini Flash & Prompt & 0\% & 65.10 & 64.45 \\
Gemini Pro & Prompt & 0\% & 63.20 & 62.85 \\ 
\hline
\midrule
Gemini Pro & ST & 1\% & 74.10 & 72.29 \\ 
Gemini Pro & MT & 1\% & \textbf{77.64} & \textbf{76.66} \\ 
\hline
Gemini Pro & ST & 10\% & \textbf{75.82} & \textbf{74.60} \\
Gemini Pro & MT & 10\% & \textbf{79.13} & \textbf{77.62}\\
\hline
Gemini Pro & ST & 100\% & 80.26 & 78.85 \\ 
Gemini Pro & MT & 100\% & \textbf{81.40} & \textbf{80.12}\\
\hline
Gemini Flash & ST & 1\% & 70.43 & 70.60 \\
Gemini Flash & MT & 1\% & \textbf{73.88} & \textbf{73.01} \\ 
\hline
Gemini Flash & ST & 10\% & 76.21 & 74.89 \\ 
Gemini Flash & MT & 10\% & \textbf{77.38} & \textbf{75.49} \\
\hline
Gemini Flash & ST & 100\% & 79.10 & 77.94 \\ 
Gemini Flash & MT & 100\% & \textbf{80.47} & \textbf{79.30} \\
\hline
\midrule
Speech-Qwen & ST & 1\% & 47.63 & 47.31 \\
Speech-Qwen & MT & 1\% & \textbf{54.54} & \textbf{53.6}0 \\
\hline
Speech-Qwen & ST & 10\% & 63.43 & 62.71 \\
Speech-Qwen & MT & 10\% & \textbf{68.27} & \textbf{67.58} \\
\hline
Speech-Qwen & ST & 100\% & 69.63 & 68.80 \\
Speech-Qwen & MT & 100\% & \textbf{71.85} & \textbf{71.09} \\
\bottomrule
\end{tabular}
\end{adjustbox}
\caption{\textbf{Comparing single-task (ST) tuning to our multi-task (MT) fine-tuning on ASK-QA's test set}.
\label{tab:askqa_results}
}
\end{table}
\subsection{Additional Experiments on Spoken-SQuAD}
In Table~\ref{tab:squad_results}, we provide our extended results on the Spoken-SQuAD corpus. 
\begin{table}[h]
\centering
\begin{adjustbox}{width=1.0\linewidth,center}
\begin{tabular}{@{}lllcc@{}}
\toprule
Base Model & App. & Data & EM $\uparrow$ & F1 $\uparrow$ \\ \midrule
FusionNet \cite{huang2017fusionnet} & -- & 100\% & 46.51 & 60.06 \\
QANet \cite{lee2019mitigating} & -- & 100\% & 49.60 & 61.85 \\
DDNet \cite{you-etal-2022-end} &  -- & 100\% & 64.10 & 77.10 \\
Whisper-Qwen & Prompt & 0\% & 59.13 & 74.08 \\
Whisper-Qwen & Prompt & 20-shot & 70.00 & 79.50 \\
Gemini Pro & Prompt & 0\% & 67.41 & 82.21 \\
\hline
\midrule
Speech-Phi & ST & 1\% & 15.08 & 25.03 \\
Speech-Phi & MT & 1\% & \textbf{22.91} & \textbf{35.02} \\
\hline
Speech-Phi & ST & 10\% & 31.43 & 44.69 \\
Speech-Phi & MT & 10\% &\textbf{ 49.32} & \textbf{63.09} \\
\hline
Speech-Phi & ST & 100\% & 50.53 & 64.46 \\
Speech-Phi & MT & 100\% & \textbf{62.14} & \textbf{74.31} \\
\midrule
\hline
Speech-Qwen & ST & 1\% & 60.25 & 73.24 \\
Speech-Qwen & MT & 1\% & \textbf{63.15} & \textbf{75.40} \\
\hline
Speech-Qwen & ST & 10\% & 62.69 & 75.94 \\
Speech-Qwen & MT & 10\% & \textbf{66.38} & \textbf{78.80} \\
\hline
Speech-Qwen & ST & 100\% & 68.75 & 80.92 \\
Speech-Qwen & MT & 100\% &\textbf{ 72.13} & \textbf{82.36}\\
\bottomrule
\end{tabular}
\end{adjustbox}
\caption{\textbf{Experimental results comparing single-task SFT (ST) and our proposed multi-task approach (MT) on Spoken SQuAD's test set}.
\label{tab:squad_results}
}
\end{table}

\paragraph{Additional Models and Baselines:} We additionally examine experiments with Speech-Phi, which we train as described in Appendix~\ref{sec:adapter_details}. This model uses Phi 3.5 Mini as the base decoder-only LLM, with up to 128k context.

We also provide the full experimental results of several baselines: FusionNet from \citet{huang2017fusionnet}, QANet from \citet{lee2019mitigating}, DDNet which is the state-of-the-art open-source model from ~\citet{you-etal-2022-end}, and Whisper-Qwen, which is a cascade-style system which uses uses Whisper-Large v3~\cite{radford2023robust} to first transcribe the audio then passes the transcription as context to Qwen 2.5 7B Instruct~\cite{yang2024qwen2} (the same model used for tuning in our experiments). We use this modular Whisper-Qwen system with both 0-shot prompting and 20-shot in-context learning. The in-context examples are given using fully textual gold transcription examples.

\paragraph{Findings:} In Table~\ref{tab:squad_results}, we consistently see that in the end-to-end speech setting, multi-task learning improves upon single-task learning. We see a particularly strong improvement using Speech-Phi. We also note that the final ability of the adapter-trained MLLM to complete the downstream SQA task may depend on the base decoder's performance on textual QA. If the projection layer is tuned to perfectly represent the audio, then the bottleneck on performance may be the decoder model's task performance on SQuAD since Spoken-SQuAD is a fully semantic task with limited acoustic diversity -- the focus in the corpus construction at the time was on discrepancies between TTS and ASR~\cite{li2018spoken}. We see that providing Qwen with golden transcripts for in-context learning in a modular system can achieve very strong performance for this very reason.

\section{Training Details}
\label{sec:training_details}
\paragraph{Open-weight models:}
Our tuning experiments using open-weight models are conducted on a single node with 8 NVIDIA A100 80GB GPUs. We rely on Deepspeed ZeRO-3~\cite{rasley2020deepspeed} and build on top of HuggingFace~\cite{wolf2020transformers}, PyTorch~\cite{paszke2019pytorch}, and SLAM-LLM~\cite{ma2024embarrassingly}. For both Speech-Qwen and Speech-Phi, we achieve our best results using an initial learning rate of 1e-4. With Speech-Qwen, we use a total batch size of 8 given our hardware constraints. For Speech-Phi, the total batch size is 16, 16, and 32 for ASK-QA, Spoken-SQuAD, and SD-QA, respectively. Our models are tuned on downstream tasks for up to 20 epochs in the limited data setting, with early stopping based on validation loss.
\paragraph{Closed-weight models:}
We perform supervised fine-tuning on ``gemini-1.5-flash-002'' and ``gemini-1.5-pro-002'' using adapters on Google Cloud's Vertex AI platform. We obtain best results using a learning rate multiplier of 1. We tune our models for a maximum of 20 epochs in the limited data setting.

\section{Risks and Ethical Considerations}
There are significant privacy concerns around speech data collection~\cite{nautsch2019preserving}, and so in this work, we rely on synthetically generated speech. However, as previously mentioned, one limitation of our work is on TTS quality. It is possible that generating long-context speech at scale will allow for hallucinations depending on the quality of the chosen TTS model. Even with automated filtering efforts, it may still be possible for these hallucinations to bypass the filtering mechanism. In our corpus, the Word Error Rate should be rather low due to the aforementioned filtering mechanism, but this still poses risk -- especially if such synthetic data are contributed to large-scale model training.

\section{Assets Used}
\label{sec:licensing}
All resources used have been cited appropriately in the paper. In this section, we enumerate each of the existing artifacts used in our work along with their license.

\textbf{Existing Models}
\begin{itemize}
    \item Gemini 1.5 Pro (gemini-1.5-pro-002), Gemini 1.5 Flash (gemini-1.5-flash-002) ~\citep{team2023gemini}: Accessed through the Google Cloud Vertex AI Platform. \url{https://cloud.google.com/products/gemini?hl=en}
    \item MiniLM-L6-v2~\citep{reimers2019sentence}: Apache 2.0. \url{https://huggingface.co/sentence-transformers/all-MiniLM-L6-v2}
    \item Qwen2.5-7B-Instruct~\citep{yang2024qwen2}: MIT Open-Source License. \url{https://huggingface.co/Qwen/Qwen2.5-7B-Instruct}
    \item WavLM~\cite{chen2022wavlm}: MIT Open-Source License. \url{https://github.com/microsoft/unilm/blob/master/wavlm/README.md}
    \item Phi-3-mini-128k-instruct~\cite{abdin2024phi}: MIT Open-Source License. \url{https://huggingface.co/microsoft/Phi-3-mini-128k-instruct}
\end{itemize}

\textbf{Existing Datasets}
\begin{itemize}
    \item Abg-CoQA~\citep{guo2021abg}: MIT Open-Source License. \url{https://github.com/MeiqiGuo/AKBC2021-Abg-CoQA}
    \item Spoken-SQuAD~\cite{li2018spoken}: Open-Source. \url{https://github.com/Chia-Hsuan-Lee/Spoken-SQuAD}
    \item SQuAD~\cite{rajpurkar2016squad}: CC-BY-SA 4.0 License. \url{https://rajpurkar.github.io/SQuAD-explorer/}
\end{itemize}

\textbf{Existing and Software}
\begin{itemize}
    \item Google Cloud Pipeline Components: Apache 2.0. \url{https://cloud.google.com/vertex-ai/docs/pipelines/components-introduction}
    \item HuggingFace Transformers~\citep{wolf-etal-2020-transformers}: Apache 2.0. \url{https://github.com/huggingface/transformers/tree/main}
    \item PyTorch~\citep{paszke2019pytorch}: PyTorch Open Source License. \url{https://github.com/pytorch/pytorch/tree/main}
    \item Vertex AI SDK: Apache 2.0. \url{https://cloud.google.com/vertex-ai/docs/python-sdk/use-vertex-ai-python-sdk}
    \item SLAM-LLM: MIT License. \url{https://github.com/X-LANCE/SLAM-LLM/tree/main}
\end{itemize}

\begin{table*}[!htbp]
\small
\begin{tabular}{p{0.05\linewidth}p{0.15\linewidth}p{0.7\linewidth}}
Usage & Modality & Content \\ \hline
Input & Speech & \textbf{Speaker 1:} A few years ago, an Englishman called Roy Jones went on holiday to a small seaside town in the west of England. He was swimming in the sea one day when, as he opened his mouth, his false teeth fell out and floated away. The following year, Mr. Jones returned to the same town. As he was having dinner in a local cafe one evening, he mentioned the story of his lost teeth to the manager. The manager looked surprised. He explained that he had found a set of false teeth on the beach last month. Then he asked Roy Jones if he wanted to try them on. OK, said Mr. Jones. I suppose it won't do any harm. When the manager brought him the teeth, Mr. Jones put them into his mouth, and laughed and laughed. They were his. In 1987, an American couple called Jane and Robert Bentley went for a picnic on a beach in California. \textbf{When they returned home, Mrs. Bentley realized that she had lost her wedding ring. It wasn't a lot of money but it was valuable to Jane Bentley. The Bentleys drove straight back to the beach, and searched for the ring for three hours, but could not find it. A few months later, Mr. Bentley went fishing off the same beach. As he pulled a large crab out of the sea, he noticed that there was something attached to one of its claws. It was his wife's wedding ring!} At the end of the 19th century, a young woman called Rose Harcourt was on her honeymoon in Barmouth, North Wales, when she lost a gold bracelet her husband had given her as a wedding gift. Feeling very upset, she went straight to the police stations and asked if anyone had found her bracelet. Unfortunately, no one had. Twenty-five years later, the Harcourts returned to Barmouth. They were sitting on the beach one day when Mrs. Harcourt noticed something gold in the sand by the edge of the sea. She walked down to see what it was, and discovered her gold bracelet that had been missing for 25 years. 

\textbf{Speaker 2:} Was it expensive? 

\textbf{Speaker 3:} No, it was not expensive. 

\textbf{Speaker 2:} Was it recovered? 

\textbf{Speaker 3:} Yes, it was recovered. 

\textbf{Speaker 2:} When?\\
Input & Text & The audio recording contains a story followed by a conversation between a User and an Assistant. You will continue the conversation for the Assistant by selecting the most appropriate response from the following: A. Do you mean the popular generic name? B. Are you asking why the dog was looking at Sue or why Jack walked up to Sue? C. More Chinese people can afford cars because of them. \textbf{D. It was recovered a few months later.} \\
\hline
Output & Text & \textbf{D. It was recovered a few months later.} \\
\bottomrule
\end{tabular}
\caption{Example of the commonsense Response Selection auxiliary task for ASK-QA.}
\label{askqa_rs}
\end{table*}

\begin{table*}[!htbp]
\small
\begin{tabular}{p{0.05\linewidth}p{0.15\linewidth}p{0.7\linewidth}}
Usage & Modality & Content \\ \hline
Input & Speech & \textbf{Speaker 1:} A few years ago, an Englishman called Roy Jones went on holiday to a small seaside town in the west of England. He was swimming in the sea one day when, as he opened his mouth, his false teeth fell out and floated away. The following year, Mr. Jones returned to the same town. As he was having dinner in a local cafe one evening, he mentioned the story of his lost teeth to the manager. The manager looked surprised. He explained that he had found a set of false teeth on the beach last month. Then he asked Roy Jones if he wanted to try them on. OK, said Mr. Jones. I suppose it won't do any harm. When the manager brought him the teeth, Mr. Jones put them into his mouth, and laughed and laughed. They were his. In 1987, an American couple called Jane and Robert Bentley went for a picnic on a beach in California. When they returned home, Mrs. Bentley realized that she had lost her wedding ring. It wasn't a lot of money but it was valuable to Jane Bentley. The Bentleys drove straight back to the beach, and searched for the ring for three hours, but could not find it. A few months later, Mr. Bentley went fishing off the same beach. As he pulled a large crab out of the sea, he noticed that there was something attached to one of its claws. It was his wife's wedding ring! At the end of the 19th century, a young woman called Rose Harcourt was on her honeymoon in Barmouth, North Wales, when she lost a gold bracelet her husband had given her as a wedding gift. Feeling very upset, she went straight to the police stations and asked if anyone had found her bracelet. Unfortunately, no one had. Twenty-five years later, the Harcourts returned to Barmouth. They were sitting on the beach one day when Mrs. Harcourt noticed something gold in the sand by the edge of the sea. She walked down to see what it was, and discovered her gold bracelet that had been missing for 25 years. 

\textbf{Speaker 2:} Was it expensive? 

\textbf{Speaker 3:} No, it was not expensive. 

\textbf{Speaker 2:} Was it recovered? 

\textbf{Speaker 3:} Yes, it was recovered. 

\textbf{Speaker 2:} When?\\
Input & Text & The audio recording contains a story followed by a conversation between a User and an Assistant. Transcribe the conversation but not the story. Provide your answer in the format 

User: [Utterance]

Assistant: [Utterance]

and so on. \\
\hline
Output & Text & \textbf{User:} Was it expensive? 

\textbf{Assistant:} No, it was not expensive. 

\textbf{User:} Was it recovered? 

\textbf{Assistant:} Yes, it was recovered. 

\textbf{User:} When?\\
\bottomrule
\end{tabular}
\caption{Example of the Dialogue Listening Comprehension auxiliary task for ASK-QA.}
\label{askqa_dlc}
\end{table*}

\begin{table*}[!htbp]
\small
\begin{tabular}{p{0.05\linewidth}p{0.15\linewidth}p{0.7\linewidth}}
Usage & Modality & Content \\ \hline
Input & Speech & \textbf{Speaker 1:} A few years ago, an Englishman called Roy Jones went on holiday to a small seaside town in the west of England. He was swimming in the sea one day when, as he opened his mouth, his false teeth fell out and floated away. The following year, Mr. Jones returned to the same town. As he was having dinner in a local cafe one evening, he mentioned the story of his lost teeth to the manager. The manager looked surprised. He explained that he had found a set of false teeth on the beach last month. Then he asked Roy Jones if he wanted to try them on. OK, said Mr. Jones. I suppose it won't do any harm. When the manager brought him the teeth, Mr. Jones put them into his mouth, and laughed and laughed. They were his. In 1987, an American couple called Jane and Robert Bentley went for a picnic on a beach in California. When they returned home, Mrs. Bentley realized that she had lost her wedding ring. It wasn't a lot of money but it was valuable to Jane Bentley. The Bentleys drove straight back to the beach, and searched for the ring for three hours, but could not find it. A few months later, Mr. Bentley went fishing off the same beach. As he pulled a large crab out of the sea, he noticed that there was something attached to one of its claws. It was his wife's wedding ring! At the end of the 19th century, a young woman called Rose Harcourt was on her honeymoon in Barmouth, North Wales, when she lost a gold bracelet her husband had given her as a wedding gift. Feeling very upset, she went straight to the police stations and asked if anyone had found her bracelet. Unfortunately, no one had. Twenty-five years later, the Harcourts returned to Barmouth. They were sitting on the beach one day when Mrs. Harcourt noticed something gold in the sand by the edge of the sea. She walked down to see what it was, and discovered her gold bracelet that had been missing for 25 years. 

\textbf{Speaker 2:} Was it expensive? 

\textbf{Speaker 3:} No, it was not expensive. 

\textbf{Speaker 2:} Was it recovered? 

\textbf{Speaker 3:} Yes, it was recovered. 

\textbf{Speaker 2:} When?\\
Input & Text & The audio recording contains a story followed by a conversation between a User and an Assistant. Transcribe the story but not the conversation.\\
\hline
Output & Text & A few years ago, an Englishman called Roy Jones went on holiday to a small seaside town in the west of England. He was swimming in the sea one day when, as he opened his mouth, his false teeth fell out and floated away. The following year, Mr. Jones returned to the same town. As he was having dinner in a local cafe one evening, he mentioned the story of his lost teeth to the manager. The manager looked surprised. He explained that he had found a set of false teeth on the beach last month. Then he asked Roy Jones if he wanted to try them on. OK, said Mr. Jones. I suppose it won't do any harm. When the manager brought him the teeth, Mr. Jones put them into his mouth, and laughed and laughed. They were his. In 1987, an American couple called Jane and Robert Bentley went for a picnic on a beach in California. When they returned home, Mrs. Bentley realized that she had lost her wedding ring. It wasn't a lot of money but it was valuable to Jane Bentley. The Bentleys drove straight back to the beach, and searched for the ring for three hours, but could not find it. A few months later, Mr. Bentley went fishing off the same beach. As he pulled a large crab out of the sea, he noticed that there was something attached to one of its claws. It was his wife's wedding ring! At the end of the 19th century, a young woman called Rose Harcourt was on her honeymoon in Barmouth, North Wales, when she lost a gold bracelet her husband had given her as a wedding gift. Feeling very upset, she went straight to the police stations and asked if anyone had found her bracelet. Unfortunately, no one had. Twenty-five years later, the Harcourts returned to Barmouth. They were sitting on the beach one day when Mrs. Harcourt noticed something gold in the sand by the edge of the sea. She walked down to see what it was, and discovered her gold bracelet that had been missing for 25 years. \\
\bottomrule
\end{tabular}
\caption{Example of the Story Listening Comprehension auxiliary task for ASK-QA.}
\label{askqa_slc}
\end{table*}

\begin{table*}[!htbp]
\small
\begin{tabular}{p{0.05\linewidth}p{0.15\linewidth}p{0.7\linewidth}}
Usage & Modality & Content \\ \hline
Input & Speech & \textbf{Speaker 1:} A few years ago, an Englishman called Roy Jones went on holiday to a small seaside town in the west of England. He was swimming in the sea one day when, as he opened his mouth, his false teeth fell out and floated away. The following year, Mr. Jones returned to the same town. As he was having dinner in a local cafe one evening, he mentioned the story of his lost teeth to the manager. The manager looked surprised. He explained that he had found a set of false teeth on the beach last month. Then he asked Roy Jones if he wanted to try them on. OK, said Mr. Jones. I suppose it won't do any harm. When the manager brought him the teeth, Mr. Jones put them into his mouth, and laughed and laughed. They were his. In 1987, an American couple called Jane and Robert Bentley went for a picnic on a beach in California. \textbf{When they returned home, Mrs. Bentley realized that she had lost her wedding ring. It wasn't a lot of money but it was valuable to Jane Bentley. The Bentleys drove straight back to the beach, and searched for the ring for three hours, but could not find it. A few months later, Mr. Bentley went fishing off the same beach. As he pulled a large crab out of the sea, he noticed that there was something attached to one of its claws. It was his wife's wedding ring!} At the end of the 19th century, a young woman called Rose Harcourt was on her honeymoon in Barmouth, North Wales, when she lost a gold bracelet her husband had given her as a wedding gift. Feeling very upset, she went straight to the police stations and asked if anyone had found her bracelet. Unfortunately, no one had. Twenty-five years later, the Harcourts returned to Barmouth. They were sitting on the beach one day when Mrs. Harcourt noticed something gold in the sand by the edge of the sea. She walked down to see what it was, and discovered her gold bracelet that had been missing for 25 years. 

\textbf{Speaker 2:} Was it expensive? 

\textbf{Speaker 3:} No, it was not expensive. 

\textbf{Speaker 2:} Was it recovered? 

\textbf{Speaker 3:} Yes, it was recovered. 

\textbf{Speaker 2:} When?\\
Input & Text & The audio recording contains a story followed by a conversation between a User and an Assistant. You will continue the conversation for the Assistant by providing the next response. If you do not have enough information or context to answer the User's last question, you should ask a clarifying question.\\
\hline
Output & Text & It was recovered a few months later. \\
\bottomrule
\end{tabular}
\caption{Example of the Response Generation task for ASK-QA.}
\label{askqa_rg}
\end{table*}
\clearpage
\begin{table*}[!htbp]
\small
\begin{tabular}{p{0.05\linewidth}p{0.15\linewidth}p{0.7\linewidth}}
Usage & Modality & Content \\ \hline
Input & Speech & Alexander Graham Bell. March 3. 1847 August 2. 1922. Was a Scottish-born. N3. Scientist. Inventor. Engineer and innovator who is credited with patenting the first practical telephone.\\

Input & Text & The audio contains a passage providing some information. Follow the User's request.

User: Transcribe the speech in the passage.\\
\hline
Output & Text & Alexander Graham Bell. March 3. 1847 August 2. 1922. Was a Scottish-born. N3. Scientist. Inventor. Engineer and innovator who is credited with patenting the first practical telephone. \\
\bottomrule
\end{tabular}
\caption{Example of the Listening Comprehension auxiliary task for Spoken-SQuAD.}
\label{squad_lc}
\end{table*}

\begin{table*}[!htbp]
\small
\begin{tabular}{p{0.05\linewidth}p{0.15\linewidth}p{0.7\linewidth}}
Usage & Modality & Content \\ \hline
Input & Speech & Alexander Graham Bell. March 3. 1847 August 2. 1922. Was a Scottish-born. N3. Scientist. Inventor. Engineer and innovator who is credited with patenting the first practical \textbf{telephone}.\\

Input & Text & The audio contains a passage providing some information. The user will ask a question about some information from the audio. The assistant should answer the user's question by selecting the correct answer choice.

User: What is Bell most famous for inventing? Choose from the following choices: A. britain \textbf{B. telephone} C. major performing arts D. london county council\\
\hline
Output & Text & \textbf{B. telephone} \\
\bottomrule
\end{tabular}
\caption{Example of the commonsense Response Selection auxiliary task for Spoken-SQuAD.}
\label{squad_rs}
\end{table*}

\begin{table*}[!htbp]
\small
\begin{tabular}{p{0.05\linewidth}p{0.15\linewidth}p{0.7\linewidth}}
Usage & Modality & Content \\ \hline
Input & Speech & Alexander Graham Bell. March 3. 1847 August 2. 1922. Was a Scottish-born. N3. Scientist. Inventor. Engineer and innovator who is credited with patenting the first practical \textbf{telephone}.\\

Input & Text & The audio contains a passage providing some information. The user will ask a question about some information from the audio. The assistant should answer the user's question using information which can be found in the passage.

User: What is Bell most famous for inventing?\\
\hline
Output & Text & \textbf{telephone} \\
\bottomrule
\end{tabular}
\caption{Example of the Response Generation auxiliary task for Spoken-SQuAD.}
\label{squad_rg}
\end{table*}

\end{document}